\documentclass[letterpaper]{article} 
\usepackage[draft]{aaai2026}  
\usepackage{times}  
\usepackage{helvet}  
\usepackage{courier}  
\usepackage[hyphens]{url}  
\usepackage{graphicx} 
\usepackage{natbib}  
\usepackage{caption} 
\usepackage{algorithm}
\usepackage{algorithmic}
\usepackage{amsmath}
\usepackage{amssymb}
\usepackage{booktabs}
\usepackage{multirow}
\usepackage{adjustbox}
\usepackage{tabularx}
\usepackage[table]{xcolor}
\usepackage{makecell}
\usepackage{threeparttable}
\usepackage{colortbl}
\usepackage{longtable}
\usepackage{array}
\usepackage{longtable}
\usepackage{newfloat}
\usepackage{listings}
\DeclareCaptionStyle{ruled}{labelfont=normalfont,labelsep=colon,strut=off} 
\floatstyle{ruled}
\newfloat{listing}{tb}{lst}{}
\floatname{listing}{Listing}
\newcommand{\benchname}{\textsc{CANVAS}}

\title{\benchname{}: A Benchmark for Vision-Language Models on Tool-Based User Interface Design}
\author{
Daeheon Jeong\textsuperscript{\rm 1}\equalcontrib,
Seoyeon Byun\textsuperscript{\rm 2}\equalcontrib,
Kihoon Son\textsuperscript{\rm 1},
Dae Hyun Kim\textsuperscript{\rm 3},
Juho Kim\textsuperscript{\rm 1}
}
\affiliations{
\textsuperscript{\rm 1}KAIST\;
\textsuperscript{\rm 2}Korea University\;
\textsuperscript{\rm 3}Yonsei University\\
\{daeheon.jeong, kihoon.son, juhokim\}@kaist.ac.kr,\; byunhwan8832@korea.ac.kr,\; dhkim16@yonsei.ac.kr
}

\begin{document}

\copyrighttext{
Preprint. This is the authors' version of a paper accepted for publication in the \textit{Proceedings of the AAAI Conference on Artificial Intelligence (AAAI-26)}.
}

\maketitle

\begin{abstract}
User interface (UI) design is an iterative process in which designers progressively refine their work with design software such as Figma or Sketch. 
Recent advances in vision language models (VLMs) with tool invocation suggest these models can operate design software to edit a UI design through iteration. 
Understanding and enhancing this capacity is important, as it highlights VLMs’ potential to collaborate with designers within conventional software.
However, as no existing benchmark evaluates tool-based design performance, the capacity remains unknown.
To address this, we introduce \benchname{}, a benchmark for VLMs on tool-based user interface design. 
Our benchmark contains 598 tool-based design tasks paired with ground-truth references sampled from 3.3K mobile UI designs across 30 function-based categories (e.g., onboarding, messaging).
In each task, a VLM updates the design step-by-step through context-based tool invocations (e.g., create a rectangle as a button background), linked to design software. 
Specifically, \benchname{} incorporates two task types: (i) \textit{design replication} evaluates the ability to reproduce a whole UI screen; (ii) \textit{design modification} evaluates the ability to modify a specific part of an existing screen.
Results suggest that leading models exhibit more strategic tool invocations, improving design quality.
Furthermore, we identify common error patterns models exhibit, guiding future work in enhancing tool-based design capabilities.
\end{abstract}

\section{Introduction}
\label{sec:introduction}
User interface (UI) design is a nonlinear process in which designers experiment with ideas through back-and-forth actions and observations~\cite{goldschmidt1991dialectics, ferreira2007agile, buxton2010sketching}.
In this process, designers continuously adjust, group, and arrange each design component, such as buttons and text, using design software (e.g., Figma and Sketch).
Working with design software is essential for designers, as they provide familiar controls over design components for efficient iterations~\cite{resnick2005design, stolterman2012design, son2024demystifying}.

\begin{figure}[t]
  \centering
  \includegraphics[width=\columnwidth]{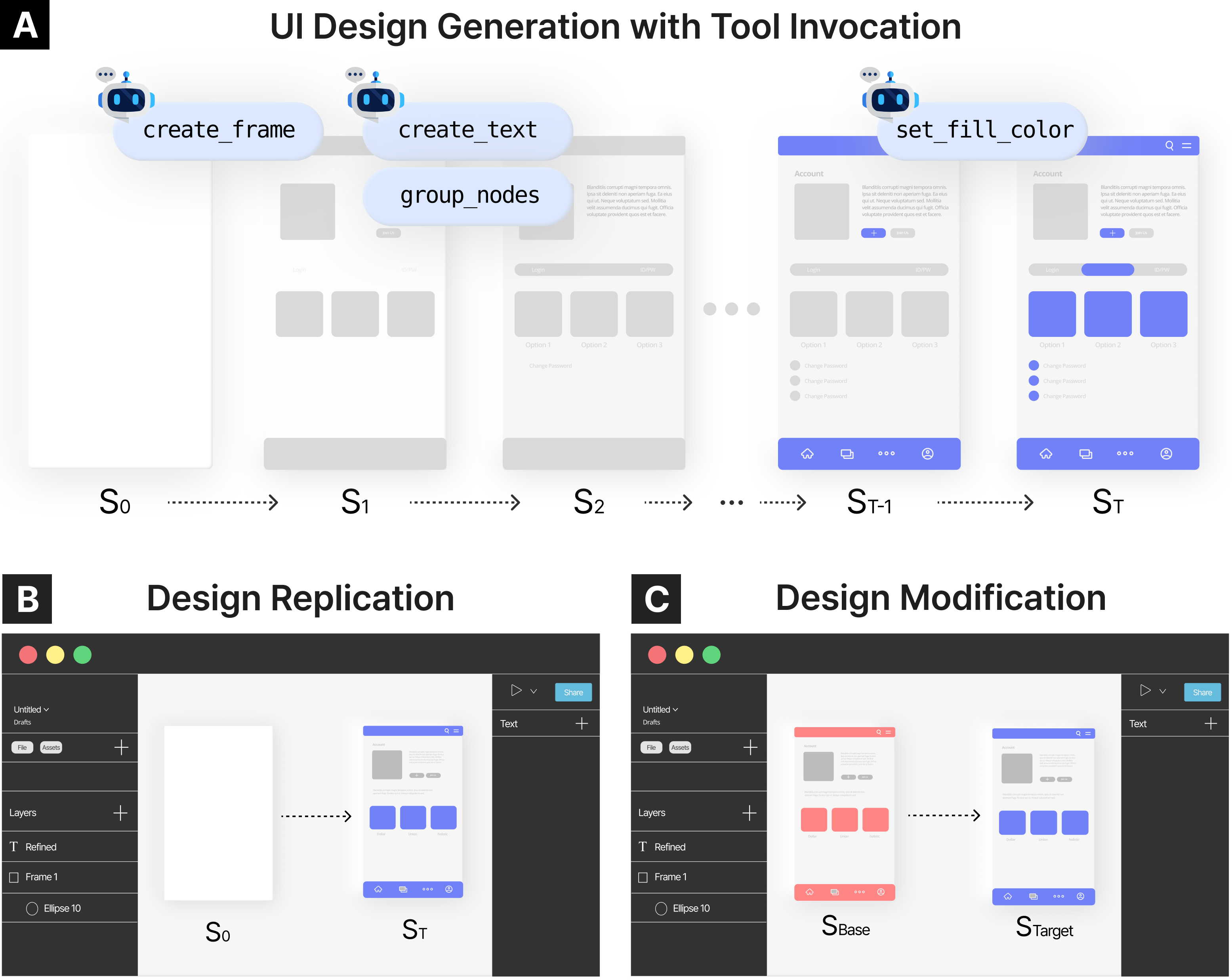}
  \caption{\textbf{Overview.} \benchname{} evaluates a VLM's capability to (A) generate a UI design with tool invocations in two tasks: (B) design replication and (C) design modification.}
  \label{fig:abstract-overview}
\end{figure}

Recent vision-language models (VLMs) have demonstrated agentic tool invocation capabilities; a model can complete tasks by invoking tools over multiple turns~\cite{niu2024screenagent}.
This capability suggests that VLMs could create and edit UI designs by interacting with design software using tool invocations (Fig.~\ref{fig:abstract-overview}A).
Yet, existing benchmarks solely focus on \textit{code-based} capabilities for the implementation stage, such as generating HTML and CSS~\cite{li2022learning, si2024design2code, li2024sketch2code}.
The earlier \textit{tool-based} design stage, where designers iteratively refine interfaces using design software, remains underexplored. 
Understanding VLMs' tool-based capabilities is crucial for enabling collaboration during this design phase~\cite{lu2024ai, choi2024creative}.

To address this gap, we introduce \benchname{}, a benchmark for evaluating VLMs' performance in tool-based UI design.
The benchmark comprises 598 UI design tasks selected to evaluate the model's performance in two task scenarios: (i) \textit{design replication} (Fig.~\ref{fig:abstract-overview}B), where the model generates a design that replicates a given UI image, and (ii) \textit{design modification} (Fig.~\ref{fig:abstract-overview}C), where the model modifies an existing human-made design by applying a sequence of edits such as resizing a component.
In each task, a VLM invokes context-based tool commands (e.g., \textit{“create a rectangle as a button background"}) to interact with the design software~\cite{figma2025figma}, and iteratively construct a complete UI design.
Each task is paired with ground-truth designs sampled from 3,327 human-crafted mobile UI designs across 30 functions (e.g., onboarding, messaging) collected from an online design community~\cite{figma2025community}.

To evaluate model performance, \benchname{} measures the similarity between generated and ground-truth designs across three hierarchical levels of visual perception, modeling how humans progressively interpret visual information: features, patterns, and objects. 
The evaluation includes structural similarity at the feature level (SSIM)~\cite{zhou2004image}, compositional similarity at the pattern level via saliency maps~\cite{jiang2023ueyes}, and semantic interpretation at the object level using BLIP captions~\cite{li2023blip}.
To capture fine-grained editing performance, we additionally evaluate component-level attributes~\cite{si2024design2code}.

In our experiments, we evaluate state-of-the-art VLMs capable of tool invocation using the \benchname{}. 
Our analysis reveals two key findings: (i) in replication tasks, model performance is closely linked to diverse and strategic tool usage; and (ii) in modification tasks, precise tool selection is critical, as a single incorrect action can cause large metric score shifts due to the granularity of the task.
Through the two task configurations, \benchname{} captures the model's capability to operate the design strategically, while precisely determining accurate tools.
Furthermore, we identify the limitations of current VLMs in tool-based design through error analysis and discuss directions for future improvement.
In summary, our contributions are:
\begin{itemize}
    \item We introduce \benchname{}, the first benchmark to evaluate VLMs' ability to perform tool-based UI generation in an interactive, multi-turn design environment.
     
    \item We construct a dataset of 598 tool-driven design tasks, comprising replication and modification tasks, sampled from 3,327 UIs across 30 categories.
    
    \item We conduct a comprehensive evaluation of five state-of-the-art VLMs and discuss key insights into their tool use behaviors through quantitative and qualitative analysis.
\end{itemize}

\section{Related Work}
\label{sec:related-work}
\subsection{UI Design Generation}

Research on UI design with generative models follows two approaches: code‑based generation, which generates the code that renders a design, and image‑based generation, which synthesizes the design image directly.
Beltramelli’s pix2code provides an early proof-of-concept for code-based UI design generation, translating UI screenshots into code using neural network models~\cite {beltramelli2018pix2code}. 
Subsequent research advances the models to condition on more abstract representations, such as sketches, wireframes, and text prompts~\cite{moran2020machine, kim2022stylette, li2024sketch2code}.
Recent work has increased instruction compliance by incorporating human and self-generated feedback into a generation process~\cite{zhou2025declarui, gui2025uicopilot}, training on the layout structure~\cite{tang2023layoutnuwa}, and setting incremental steps~\cite{wan2024automatically}.
Another line of research explored directly generating UI designs as images, including generative adversarial networks~\cite{li2019layoutgan, zhao2021guigan} and diffusion models~\cite{cheng2023play, garg2025controllable}.
These approaches are unconstrained by code-specific syntax and render design and layout images through a denoising process.
Nevertheless, the generated result remains less editable than designs created with design software, restricting its practical applicability; the problem persists in code-based generation~\cite{yuan2024towards, yuan2024maxprototyper}.
Contrary to previous approaches, \benchname{} evaluates tool-based design generation: models produce designs with tool invocations in design software, creating directly editable outputs.

\subsection{UI Design Datasets and Benchmark}

Existing datasets and benchmarks on UI generation primarily focus on code-based generation.
A common dataset structure pairs UI screenshots with corresponding source code (e.g., HTML, XML) to support code-based design generation~\cite{si2024design2code, yun2024web2code}.
The datasets have increased the scale, starting from RICO on mobile UI design~\cite{deka2017rico} to large-scale real-world datasets, such as WebCode2M~\cite{gui2025webcode2m} and WebSight~\cite{laurenccon2024unlocking}. 
These datasets enable training UI generation models and benchmarking by comparing generated images and source code with ground-truth designs. 
Evaluation metrics include pixel-level similarity, SSIM for structural comparison, CLIP for semantic alignment, and FID~\cite{zhao2021guigan, xiao2024interaction2code, li2024sketch2code, duan2025automated, gui2025uicopilot}.
Studies also utilize the source code to measure component-level attributes (e.g., text, position, color)~\cite{si2024design2code, gui2025uicopilot} and structural relationships~\cite {gui2025webcode2m}.
Still, existing metrics have limited applicability to tool-based UI design, as design software structures data representation around designer-centric concepts such as masks and layers.
\benchname{} addresses these limitations by proposing metrics adapted to the tool-based design context.

\section{Benchmark Design}
\label{sec:benchmark-design}
\subsection{Task Design}

To reflect real-world user scenarios, \benchname{} consists of two types of user interface (UI) design tasks: (i) \emph{design replication} and (ii) \emph{design modification}.
In \emph{design replication}, the model completes a design based on a reference image and a build instruction, which provides the task context and the overall design goal.
In \emph{design modification}, the model refines an existing design based on a reference image and an edit instruction, which describes the target component and the expected changes in plain language without numeric values.
These task settings reflect the designers’ practical expectations towards AI assistance, ranging from automating repetitive tasks to assisting with micro-level design edits~\cite{lu2022bridging, li2024user}.

\benchname{} represents a UI design as a state composed of components (e.g., buttons, text) and their attributes (e.g., position, width, color), mirroring how design software organizes design components.
In each task, a UI design is formalized as a state \(s\) made up of \(n\) component–attribute pairs.
\[
s \;=\; \{\, (c_i,\,a_{ij},\,v) \mid i=1,\dots,n,\; j=1,\dots,m_i \,\}.
\]

where each component \(c_i\) is associated with a set of attribute-value pairs \((a_{ij},v)\).
The benchmark evaluates the model-generated design \(\hat{s}\) by comparing it with a ground-truth design \(s_{\mathrm{GT}}\), using these attribute values and visual characteristics.

\subsubsection{Design Replication}
The replication task measures a VLM's ability to map out a sequence of tool invocations to reconstruct a target UI design \(s_{\mathrm{GT}}\).
Starting from an empty canvas \(s_0=\varnothing\), the model iteratively adds or edits components, producing a design trajectory  
\[
s_0 \;\rightarrow\; s_1 \;\rightarrow\; \dots \;\rightarrow\; s_t.
\]  
The process stops when the model determines \(s_t \approx s_{\mathrm{GT}}\) and ends tool invocations.

\subsubsection{Design Modification}
\begin{figure}[t]
  \centering
  \includegraphics[width=\columnwidth]{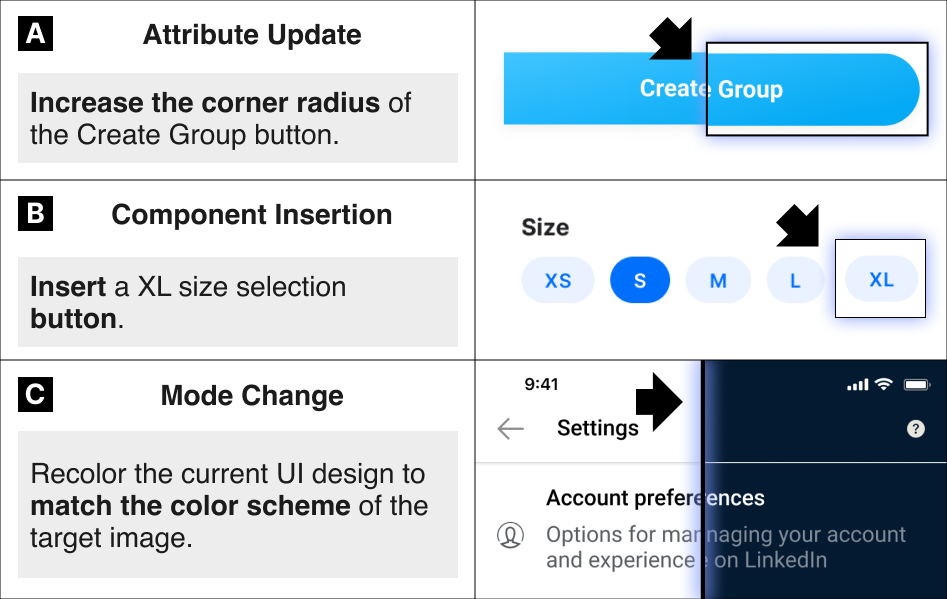}
  \caption{\textbf{Design Modification Tasks.} Design modification task measures a model’s capacity to perform targeted edits, requiring it to (A) adjust a component’s attributes, (B) insert a new component, or (C) switch the overall color scheme.}
  \label{fig:tasks}
\end{figure}

The design modification task (Fig.~\ref{fig:tasks}) measures the VLM's ability to apply a fine-grained tool invocation over turns to complete specified changes.
Starting from a pre-existing design state \(s_{old}\), the model is instructed to modify the design into a specified target state \(s_{GT}\) based on (i) a task instruction and (ii) a ground truth image.
The model needs to conduct a list of changes \(\Delta=[\dots]\) to reach the target design state 
\[
s_{old}\xrightarrow{\;\Delta\;}s_{GT}
\]
The design modification task consists of three sub-tasks:

\begin{itemize}
    \item \textit{Attribute Update} (Fig.~\ref{fig:tasks}B-(A)): Modify attributes \(a\) in different components \(c\) to target values \(v'\) in a design.  
    The attributes include fill color, size, text content, corner radius, and position.
    \[
    \Delta_{\text{attr}}
      = \{\, (c_k,\,a_k,\,v_k') \mid k=1,\dots,K \,\}.
    \]
    \item \textit{Component Insertion} (Fig.~\ref{fig:tasks}B-(B)): Create and insert a list of new component-attribute pairs into a design.
    \[
    \Delta_{\text{add}}
    = \{\, (c_k,\,a_k,\,v_k) \mid k = 1,\dots,K \,\}.
    \]
    \item \textit{Mode Change} (Fig.~\ref{fig:tasks}B-(C)): Change a list of colors to a target value to convert the design between light and dark themes.
    \[
    \Delta_{\text{col}}
      = \{\, (\mathrm{rgb}^{\text{old}}_k,\;\mathrm{rgb}^{\text{new}}_k) \mid k = 1,\dots,K \,\}.
    \]
\end{itemize}

These operations cover the primitive modification patterns in UI design workflows: property adjustments, component additions, and systematic style changes.

\subsection{Dataset Creation}
We collected UI designs from an online community and refined them through a designer review.

\subsubsection{Data Source}
Our dataset comprises UI designs from public projects on the Figma Community platform~\cite{figma2025community}, a widely used online repository.
We selected Figma as a data source for its popularity and standardized format~\footnote{All designs are licensed under Creative Commons (CC BY 4.0), and the dataset includes a link to the original project.}.

\begin{figure*}[!t]
  \centering
  \includegraphics[width=\textwidth]{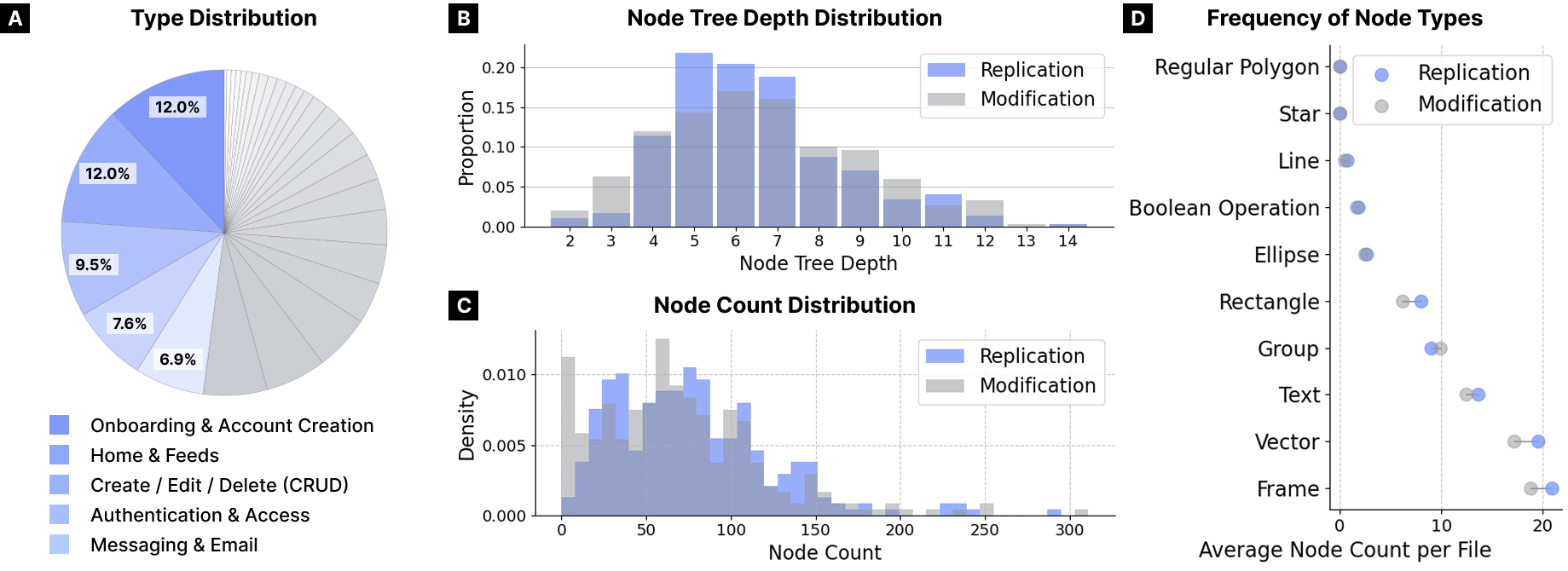} 
  \caption{\textbf{\benchname{} Data Statistics}:  (A) Frequency of the five most frequent UI design types, with other categories grouped (see Appendix for full distribution). (B) The distribution of node tree depth is similar across the replication and modification sets with a Gaussian-like pattern. (C) The node count distribution is also similar across both sets. (D) The skewed frequency of node types per design indicates common patterns in component usage.}  
  \label{fig:datacharacteristics}
\end{figure*}

\subsubsection{Data Collection}
We created the dataset through selection, categorization, sampling, and annotation to ensure UI type diversity (see Appendix for details).

\begin{itemize}
\item{(i) \emph{Data Selection}.} 
We manually selected projects with editable UI designs, excluding non-UI resources, e.g., prototyping kits. 
Selection stopped when additional designs showed minimal variation. 
Each design was exported as SVG, PNG, and JSON via Figma's REST API.

\item{(ii) \emph{Data Sampling}.} 
The designs were categorized into 30 UI types by GPT-4.1-Mini~\cite{openai2025gpt41mini} with zero-shot prompting; one of the authors defined the types, adapting Mobbin’s categorization with GPT-4.5~\cite{mobbin2025mobbin, openai2025gpt45}. 
The replication set $(n=298)$ was sampled using stratified sampling from the original pool $(n=3,327)$, covering all UI categories (10 per category, except one with 8).
The modification set $(n = 300)$ was manually sampled from the same pool based on task-relevant attributes (e.g., round borders).

\item{(iii) \emph{Data Annotation}.}
For attribute update and component insertion tasks, we generated instructions using GPT-4.1-Mini based on visual differences between manually created ``base'' and ``target'' states.
These states were manually created by editing the original design in Figma (e.g., deleting a component).

\end{itemize}

\subsubsection{Data Refinement}

To improve dataset quality, we reviewed and refined the dataset through designer review of designs and annotations. (see Appendix for details).

\begin{itemize}
\item{(i) \emph{Data Analysis}.}
We cleaned the designs through automated pre-processing (e.g., placeholder replacement, font standardization) and defined common issues, including occlusions, illustrations, and visual clutter (misleading layout composition).

\item{(ii) \emph{Manual Revision}.}
Four experienced designers revised 598 designs by fixing issues, correcting flawed instructions from GPT-4.1-Mini, and annotating required tools. 
This process resolved occlusions in 33.4\% of the designs, illustration errors in 15.5\%, and clutter in 29.0\%.
\end{itemize}

\subsection{Data Statistics}

\subsubsection{Overview}
\benchname{} contains 598 design tasks collected from 3,327 mobile UI designs across 30 categories collected through stratified sampling (Fig.\ref{fig:datacharacteristics}A). 
Each design includes an image and a JSON file representing Figma’s node structure, where components (e.g., vector, text, rectangle) are organized in hierarchical trees with attributes such as position, size, and color (see Appendix for an example). 
Vector, text, and rectangle nodes are most common, while group and frame nodes also appear frequently (Fig.\ref{fig:datacharacteristics}D).
Through the data refinement process, we remove outliers with an excessive number of nodes, reducing variance within the dataset.

\subsubsection{Structural Characteristics}

We analyzed the structural properties of our benchmark tasks to ensure they offer an appropriate level of complexity for evaluating tool-based design capabilities. 
First, the node tree depth is slightly shallower in both replication ($M=6.48,\ SD=2.04$) and modification ($M=6.62,\ SD=2.40$) tasks than in the source dataset ($M=7.06,\ SD=2.32$) (Fig.~\ref{fig:datacharacteristics}B); the node count is also lower (replication: $M=76.30$, modification: $M=69.11$, source: $M=126.25$; Fig.~\ref{fig:datacharacteristics}C), achieving a more efficient representation of the design while preserving core structure.
Despite this reduction, the normalized \textit{Shannon entropy} of the node types remains comparable, with $J = 0.763 \; (K = 10)$ for replication, $J = 0.760 \; (K = 10)$ for modification, and $J = 0.726 \; (K = 13)$ for source dataset, retaining component type diversity in the design.
The distribution of component types remains consistent across tasks, with nodes such as \texttt{frame}, \texttt{text}, and \texttt{vector} most frequent (Fig.~\ref{fig:datacharacteristics}D). 
These patterns suggest that our benchmark tasks simplify structural complexity while maintaining component-level information, enabling consistent evaluation of model performance.

\subsection{Evaluation Metrics}

Our evaluation framework measures perceptual and semantic alignment between generated and reference designs based on human visual processing. 
The framework decomposes visual similarity into three hierarchical levels that approximate the bottom-up stages of human visual perception: \textit{features} (low-level properties such as edges and colors), \textit{patterns} (mid-level compositions such as shapes and groupings), and \textit{semantic objects} (high-level interpretations such as buttons or input fields)~\cite{ware2010visual}. 
This hierarchy reflects how humans sequentially extract meaning from a design, progressively assembling basic visual features into recognizable patterns and eventually inferring functional design elements. 
Additionally, to capture model performance on component-level attribute editing (position, color, text), we include component-wise similarity, which evaluates how accurately individual UI elements are reproduced (see Appendix for implementation).

\definecolor{best}{HTML}{B8C7FF}

\begin{table*}[!ht]
\centering
\caption{\textbf{\benchname{} scores on design replication and modification task.} Each score indicates the average similarity scores between completed designs and the ground truth. In replication, GPT-4.1 and Gemini-2.5-Pro exhibit leading performance in the replication task, while GPT-4.1 consistently exhibits robust performance in the modification task.}
\label{tab:main-result}
\begin{adjustbox}{max width=\textwidth}
\begin{tabular}{l|cccc|cccc}
\toprule
\multirow{2}{*}{\textbf{Model}} &
\multicolumn{4}{c|}{\textbf{Replication}} &
\multicolumn{4}{c}{\textbf{Modification}} \\
\cmidrule(r){2-5} \cmidrule(r){6-9}
& \textbf{SSIM} & \textbf{Saliency} & \textbf{BLIP} & \textbf{Comp.\ Wise} &
  \textbf{SSIM} & \textbf{Saliency} & \textbf{BLIP} & \textbf{Comp.\ Wise} \\
\midrule
GPT-4o              & 0.739 (±0.172) & 0.478 (±0.136) & 0.495 (±0.250) & 0.671 (±0.087) &
                     0.843 ($\Delta 0.136$) & 0.845 ($-\Delta 0.009$) &
                     0.740 ($-\Delta 0.021$) & 0.943 ($\Delta 0.015$) \\
GPT-4.1             & 0.767 (±0.129) & 0.612 (±0.137) &
                     \cellcolor{best}\textbf{0.655 (±0.251)} &
                     \cellcolor{best}\textbf{0.716 (±0.075)} &
                     \cellcolor{best}\textbf{0.890 ($\Delta 0.183$)} &
                     \cellcolor{best}\textbf{0.861 ($\Delta 0.007$)} &
                     \cellcolor{best}\textbf{0.806 ($\Delta 0.044$)} &
                     \cellcolor{best}\textbf{0.951 ($\Delta 0.024$)} \\
Claude-3.5-Sonnet   & 0.725 (±0.180) & 0.483 (±0.151) & 0.518 (±0.272) & 0.666 (±0.089) &
                     0.816 ($\Delta 0.109$) & 0.858 ($\Delta 0.004$) &
                     0.775 ($\Delta 0.013$) & 0.946 ($\Delta 0.018$) \\
Gemini-2.5-Flash    & 0.736 (±0.184) & 0.619 (±0.149) & 0.571 (±0.270) & 0.702 (±0.100) &
                     0.874 ($\Delta 0.167$) & 0.857 ($\Delta 0.003$) &
                     0.784 ($\Delta 0.023$) & 0.948 ($\Delta 0.020$) \\
Gemini-2.5-Pro      & \cellcolor{best}\textbf{0.774 (±0.117)} &
                     \cellcolor{best}\textbf{0.630 (±0.162)} &
                     0.620 (±0.273) & 0.694 (±0.094) &
                     0.867 ($\Delta 0.159$) & 0.851 ($-\Delta 0.003$) &
                     0.804 ($\Delta 0.043$) & 0.935 ($\Delta 0.007$) \\
\bottomrule
\end{tabular}
\end{adjustbox}
    \begin{tablenotes}[para,flushleft]\footnotesize
      ± indicates the standard deviation; $\Delta$ indicates the average score increase from the base design; \textcolor{best}{\rule{6pt}{6pt}} marks the best score
      in each column.
    \end{tablenotes}
\end{table*}

\paragraph{Structural Similarity Index Measure} (\textit{Feature-level}): SSIM compares local image statistics (mean, variance, and covariance) to assess how well low-level features such as size and shape are preserved~\cite{zhou2004image}.

\paragraph{Saliency Similarity} (\textit{Pattern-level}): 
Saliency similarity measures predicted human attention patterns within a design. We compute histogram intersection between normalized saliency maps of the ground truth and generated design using UEyes~\cite{jiang2023ueyes}, trained on eye-tracking data from UI screens.

\paragraph{BLIP Caption Similarity} (\textit{Object-level}):
We generate captions for both ground-truth and generated designs using BLIP-2 and compute cosine similarity between their embeddings using SentenceTransformer~\cite{reimers2019sentence}, approximating semantic-level similarity.

\subsubsection{Component-wise Similarity}

Following Design2Code~\cite{si2024design2code} and related work~\cite{li2024sketch2code}, we perform one-to-one component matching using the Hungarian algorithm (IoU for visual elements, text-position similarity for text), then compute similarity across four attributes: component match rate, position (Euclidean distance), text content (F1 score), and fill color (RGB distance).

\subsubsection{Metric Aggregation}

For both tasks, we evaluate the similarity between the generated design $s_t$ and the reference design $s_{GT}$, reporting the final score as the average similarity values across all benchmark cases. 
In the modification task, we measure how much the model's edits improve similarity to the ground truth $s_{GT}$ by comparing scores before and after editing.

\section{Experiments}
\label{sec:experiments}
\subsection{Configuration}

\subsubsection{Pipeline}

\benchname{} implements a tool-based UI design generation pipeline based on the Model Context Protocol (MCP)\footnote{Architecture building upon https://github.com/grab/cursor-talk-to-figma-mcp, MIT License}. 
Models have access to 50 predefined tools for design operations, including creation, deletion, layout adjustment, styling, and content modification, which are relayed to Figma via MCP (See Appendix for details).
Models initiate each task by receiving the target screen UI image and task instructions, employing predefined tools; for modification tasks specifically, the pipeline initializes existing designs on the Figma canvas from a structured JSON representation.
In each task, models operate within an agentic loop based on the ReAct framework~\cite{yao2023react}, which involves cycles of thought, action (tool invocation), and observation.

\subsubsection{Model Setup}

We evaluate five state-of-the-art vision-language models (VLMs) with visual understanding and tool-use capabilities, including GPT-4o~\cite{openai2024gpt4o}, GPT-4.1~\cite{openai2025gpt41}, Claude-3.5-Sonnet~\cite{anthropic2024claude35sonnet}, Gemini-2.5-Flash and Pro~\cite{comanici2025gemini}. 
All models are instructed using standardized instructions with the temperature level at 0 for reproducibility (See Appendix for the instructions). 
We excluded open-source models due to their high failure rates in multi-turn tool invocation. 
These models commonly terminated after one or two turns, resulting in blank or incomplete outputs.

\subsection{Main Results}

Table~\ref{tab:main-result} shows average similarity scores between completed designs and ground truth across the replication and modification tasks. 
The score with (±) represents the standard deviation in the replication task, while the score with ($\Delta$) indicates score changes due to model actions from the initial state $s_{old}$ to the target state $s_{GT}$ in the modification task.

\subsubsection{Replication}

As shown in Table~\ref{tab:main-result}, Gemini-2.5-Pro scored highest in SSIM and saliency, indicating strength in feature-to-pattern-level similarity (contours and shape compositions). 
GPT-4.1 achieved the best scores for BLIP and component-wise similarity, indicating strength in replicating design semantics and preserving component attributes (position, color, text)

\subsubsection{Modification}

GPT-4.1 achieved the highest scores across all four metrics. Importantly, several models demonstrate small or negative $\Delta$ values, indicating score stagnation or degradation due to model actions. We explore these results in the analysis section.

\subsection{Analysis}

To understand performance differences across models, we analyzed tool invocation histories at each turn. The histories directly capture model actions and their influence on overall scores. Our analysis reveals two primary observations: (i) models achieving higher scores in the replication task demonstrate diverse tool invocations over increased turns; and (ii) models performing well in the modification task exhibit precise, targeted tool invocations within fewer turns.

\subsubsection{Replication task demands diverse invocation of tools.}

\begin{figure}[h]
  \centering
  \includegraphics[width=\columnwidth]{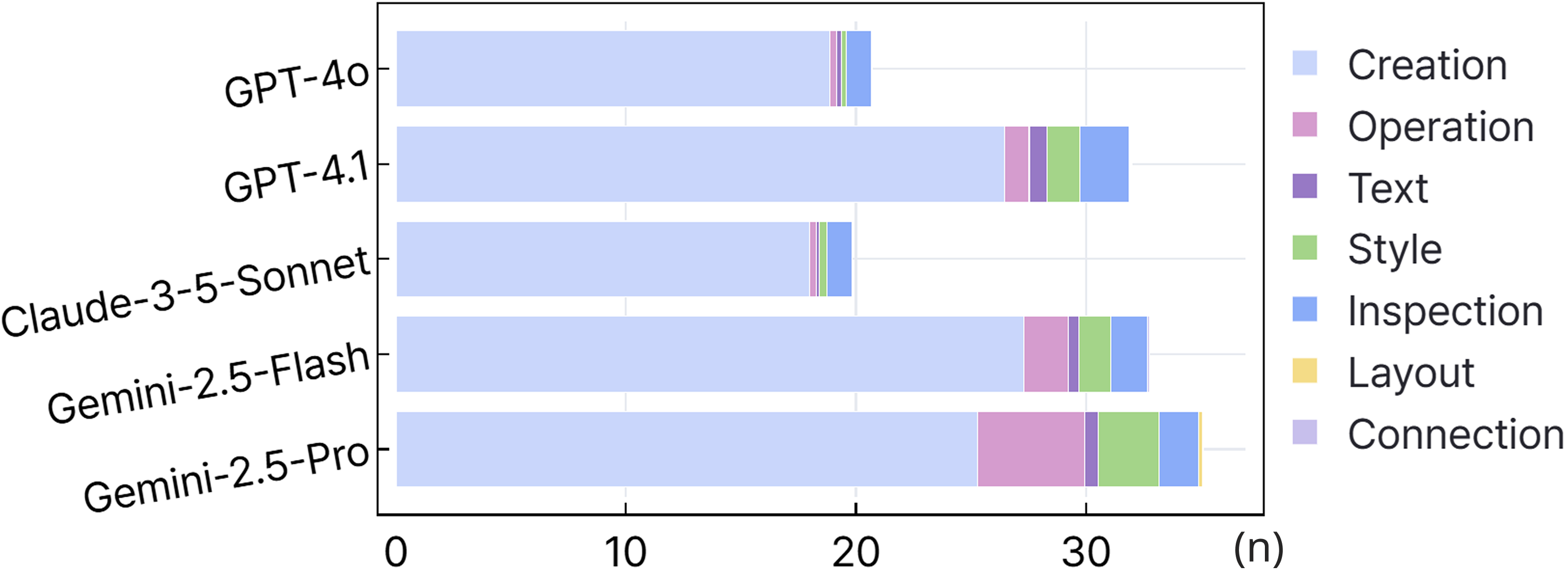}
  \caption{\textbf{Tool Invocation Frequency}: Average tool invocations per task $x \,axis$ ($n$). Colored blocks show tool types (e.g., creation includes \texttt{create\_rectangle}). Higher-performing models exhibit greater tool diversity.}
  \label{fig:tool-distribution}
\end{figure}

We identify that in the replication task, high-scoring models exhibit more diverse tool patterns. 
Figure~\ref{fig:tool-distribution} shows average tool invocation frequency and tool type diversity by model on each task.
The higher-performing models, namely GPT-4.1 and Gemini-2.5-Pro, display increased tool diversity, indicating more strategic behaviors.
For example, we observe a modular approach, where a model creates a component (e.g., a buy button) once and propagates it across the design using the \texttt{copy\_node} action, saving turns.
In contrast, lower-performing models, such as GPT-4o and Claude-3.5-Sonnet, complete the design by simply creating components in order.
These observations suggest that intrinsic rewards encouraging exploration could help models learn strategic, diverse tool patterns.

\subsubsection{Modification task requires precise selection of tools.}

We posit that the modification task poses a unique challenge for models due to its high granularity. 
It requires precise tool invocations, where slight inaccuracies lead to non-uniform impacts across similarity metrics. 
For instance, adding a line break may minimally affect textual content but can cause a large shift in a saliency map. 
This phenomenon is reflected in the small or negative $\Delta$ values in Table~\ref{tab:main-result} (Modification); due to non-uniform shifts, the score changes converge towards zero on average.

For a more in-depth understanding, we introduce Pos@K, a metric counting the number of cases where the number (K) of similarity scores increases after an edit. 
For instance, A Pos@3 indicates the number of cases where the model edit introduced a positive increase in \emph{three} metrics and a decrease in the remaining metrics. 
Table~\ref{tab:passk-singlecol} shows the distribution of Pos@K values for each model in the modification Task 1, \textit{attribute update}, in ratio. 
Notably, in the table, models that underperformed elsewhere --- Claude-3.5-Sonnet and Gemini-2.5-Flash --- yielded a higher proportion of Pos@$4$ than other models.

\definecolor{greenblock}{HTML}{E2F4E2}

\newcommand{\bestgreen}[1]{\cellcolor{greenblock}{\textcolor{black}{\textbf{#1}}}}

\begin{table}[h]
\centering
\caption{Distribution of Pos@K cases in ratio for each model in Task 1: Attribute Update.}
\label{tab:passk-singlecol}
\begin{adjustbox}{max width=\columnwidth}
\begin{tabular}{lcccccc}
\toprule
\textbf{Model} & \textbf{Pos@4} & \textbf{Pos@3} & \textbf{Pos@2} & \textbf{Pos@1} & \textbf{Pos@0} & \textbf{F} \\
\midrule
GPT-4o              & 24.0\% & \bestgreen{41.0\%} & 18.0\% & 13.0\% & 2.0\% & 2.0\% \\
GPT-4.1             & 25.0\% & 31.0\% & \bestgreen{23.0}\% & 14.0\% & 6.0\% & 1.0\% \\
Claude-3.5-Sonnet   & 30.0\% & 38.0\% & 17.0\% & 10.0\% & 4.0\% & 1.0\% \\
Gemini-2.5-Flash    & \bestgreen{31.0\%} & 37.0\% & 20.0\% & 9.0\% & 3.0\% & 0.0\% \\
Gemini-2.5-Pro      & 27.0\% & 36.0\% & 12.0\% & \bestgreen{16.0}\% & \bestgreen{8.0}\% & 1.0\% \\
\bottomrule
\end{tabular}
\end{adjustbox}
    \begin{tablenotes}[flushleft]
      \footnotesize
      \item \textcolor{greenblock}{\rule{6pt}{6pt}}: the highest value in each column. F: the ratio of failed cases.
    \end{tablenotes}
\end{table}

To investigate this performance inversion, we compared tools invoked by models against human-annotated tools for each task. 
Three designers annotated the required tools for each of the 300 modification cases, serving as reference annotations (see Appendix for details). 
As shown in Table~\ref{tab:tool-invocation}, Gemini-2.5-Pro and GPT-4.1 exhibited lower tool precision but higher diversity compared to Claude-3.5-Sonnet and Gemini-2.5-Flash.
This suggests that excessive tool diversity impaired performance in this task.
The finding generalizes across all modification tasks: treating Pos@K as an ordinal score, we found a positive correlation between tool precision and the Pos@K score $(\rho\ = 0.149, p<0.01)$ and a negative correlation between tool diversity and Pos@K $(\rho=-0.365, p<0.01)$.
These results indicate that precise tool selection is critical for the modification task.
To improve precision, methods that teach accurate skills, such as imitation learning, are necessary.

\begin{table}[t]
\centering
\small
\setlength\tabcolsep{6pt}
\caption{Tool invocation statistics per model.}
\label{tab:tool-invocation}

\begin{adjustbox}{width=\columnwidth}
\begin{tabular}{lccc}
\toprule
\textbf{Model} & \textbf{Tool Precision} & \textbf{Tool Recall} & \textbf{Tool Diversity} \\
\midrule
GPT-4o               & 0.4195 & 0.9967 & 4.3100 \\

\rowcolor{gray!15}%
GPT-4.1              & 0.5434 & 1.0000 & 5.4433 \\

Claude-3.5-Sonnet    & 0.5677 & 0.9900 & 3.6300 \\
Gemini-2.5-Flash     & 0.5943 & 0.9767 & 3.4067 \\

\rowcolor{gray!15}%
Gemini-2.5-Pro       & 0.5659 & 0.9633 & 4.1400 \\
\bottomrule
\end{tabular}
\end{adjustbox}

\begin{tablenotes}[flushleft]
\footnotesize
\item Tool Precision and Recall: normalized intersection with human-annotated tools. Tool Diversity: average unique tools per case.
\end{tablenotes}

\end{table}

\subsubsection{Our metrics align with design experts.}

A study with human annotators reveals that our metrics closely align with human preference.
Table~\ref{tab:human-preference-regression} demonstrates that three metrics, saliency, BLIP, and component-wise similarity, are statistically significant predictors of pairwise human judgments on design similarity.

Following the methodology of Design2Code~\cite{si2024design2code}, we collected pairwise human preference data from design experts on Prolific. 
We used 100 design cases, chosen from our replication results via stratified sampling (weighted by UI type and complexity).
Each pair received ``win,'' ``lose,'' or ``tie'' labels from three annotators, with final outcomes determined by majority vote, excluding cases without a majority.
Subsequently, we trained a logistic regression model to predict the binary preference (win=1, lose=0) using the difference in our metric scores between two designs in a pair as the independent variable. 
On a 50/50 training/test split of the 363 non-tie samples, the model achieved 75\% prediction accuracy.

\begin{table}[h]             
\small                       
\centering
\caption{Logistic regression on human pairwise preferences using similarity metrics as features.}
\begin{tabular}{lrrrr}
\hline
            & \textbf{coef} & \textbf{std err} & \textbf{z} & \textbf{p} \\
\hline
SSIM                      & -0.0500 & 0.2241 & -0.223 & 0.82336 \\
Saliency             & 0.9808 & 0.2758 &  3.557 & 0.00038 \\
BLIP &  0.7323 & 0.2357 & 3.106 & 0.00189 \\
Comp. Wise            & 0.5836 & 0.2806 &  2.080 & 0.03506 \\
\hline
\end{tabular}
\label{tab:human-preference-regression}
\end{table}

\subsection{Ablation Study}

We conducted an ablation study to isolate the effects of multi-turn iteration and tool-based interaction.
(i) The baseline setting is tool-based multi-turn, which is identical to the setting in the \benchname{} benchmark. 
To control the effect of the turns, we include the (ii) tool-based single-turn condition, where all tool invocations are generated at once without intermediate feedback.
This enables a direct comparison against our third condition, (iii) a code-based single-turn adapted from prior work~\cite{si2024design2code}, thereby isolating the impact of using tools.

\definecolor{bestoverall}{HTML}{E8E8E8}

\begin{table}[h]
\centering
\caption{Ablation study results for each condition.}
\label{tab:ablation-study-result}

\begin{adjustbox}{max width=\columnwidth}
\begin{tabular}{ll|ccc}
\toprule
\multirow{2}{*}{\textbf{Method}} & \multirow{2}{*}{\textbf{Model}} &
\multicolumn{3}{c}{\textbf{Replication}} \\
\cmidrule(r){3-5}
& & \textbf{SSIM} & \textbf{Saliency} & \textbf{BLIP} \\
\midrule
\multirow{2}{*}{\makecell[l]{Code\\(Single-Turn)}}
& GPT-4.1        & 0.773 (±0.110) & 0.649 (±0.129) & \cellcolor{bestoverall}\textbf{0.689 (±0.231)} \\
& Gemini-2.5-Pro & 0.771 (±0.108) & 0.696 (±0.120) & 0.685 (±0.233) \\
\midrule
\multirow{2}{*}{\makecell[l]{Tool\\(Single-Turn)}}
& GPT-4.1        & 0.704 (±0.219) & 0.510 (±0.156) & 0.534 (±0.281) \\
& Gemini-2.5-Pro & \cellcolor{bestoverall}\textbf{0.775 (±0.118)} & \cellcolor{bestoverall}\textbf{0.699 (±0.121)} & 0.660 (±0.259) \\
\midrule
\multirow{2}{*}{\makecell[l]{Tool\\(Agent)}}
& GPT-4.1        & 0.767 (±0.129) & 0.612 (±0.137) & 0.655 (±0.251) \\
& Gemini-2.5-Pro & 0.774 (±0.117) & 0.630 (±0.162) & 0.620 (±0.273) \\
\bottomrule
\end{tabular}
\end{adjustbox}

\begin{tablenotes}[flushleft]
\footnotesize
\item \textcolor{bestoverall}{\rule{6pt}{6pt}}: the \emph{best overall} in each column.
\end{tablenotes}

\end{table}

Table~\ref{tab:ablation-study-result} shows that Gemini-2.5-Pro achieves its highest scores in the single-turn tool-based setting, while GPT-4.1 performs better under the multi-turn condition. 
The single-turn and multi-turn settings induce distinct model behaviors: the multi-turn setting encourages strategic, non-linear operations (e.g., layer reordering), while the single-turn setting favors more uniform, linear action sequences.
This divergence explains the performance results in Table~\ref{tab:ablation-study-result}: the riskier strategic operations (e.g., copying incorrectly designed buttons) can degrade output quality, leading to lower scores in some design cases. 
Nevertheless, we propose that this multi-turn, agentic configuration holds important potential for complex, collaborative design tasks.

\subsection{Error Cases}
We conducted an error analysis to identify common failure cases of current models and their underlying causes.

\subsubsection{Geometric Operation Errors}

\begin{figure}[H]
  \centering
  \includegraphics[width=\columnwidth]{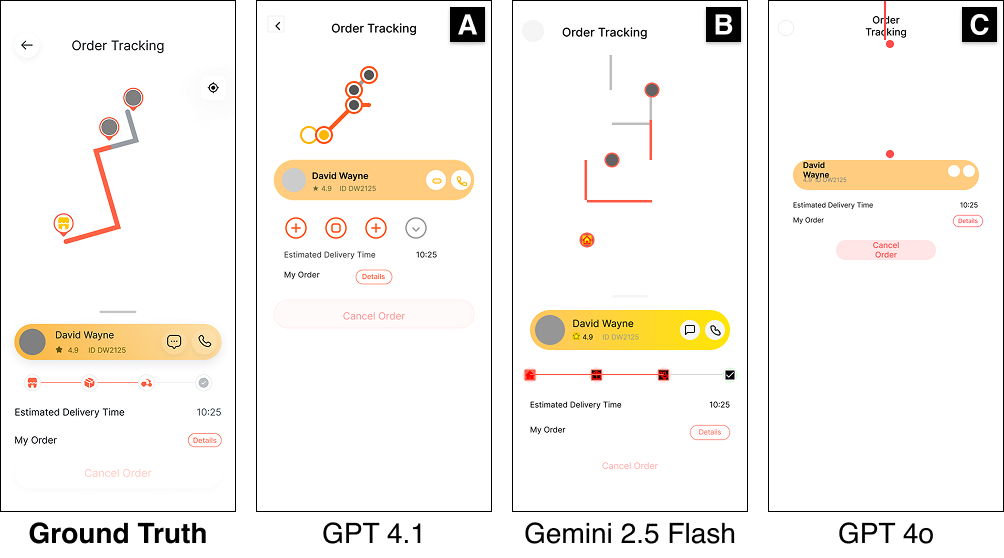}
  \caption{\textbf{Error case 1}. The models (A) miscount the markers, (B) draw irregular lines, and (C) create an inconsistent layout.}
  \label{fig:error-type-position}
\end{figure}

Models frequently fail to control the geometric properties of UI elements, leading to errors in count, directionality, and spatial arrangement. 
For instance, in the map user interface shown in Fig.~\ref{fig:error-type-position}, the model produces (A) an incorrect number of location markers, (B) paths with incoherent directions, and (C) a spatially inconsistent arrangement of components.

\subsubsection{Layout Operation Errors}

\begin{figure}[H]
  \centering
  \includegraphics[width=\columnwidth]{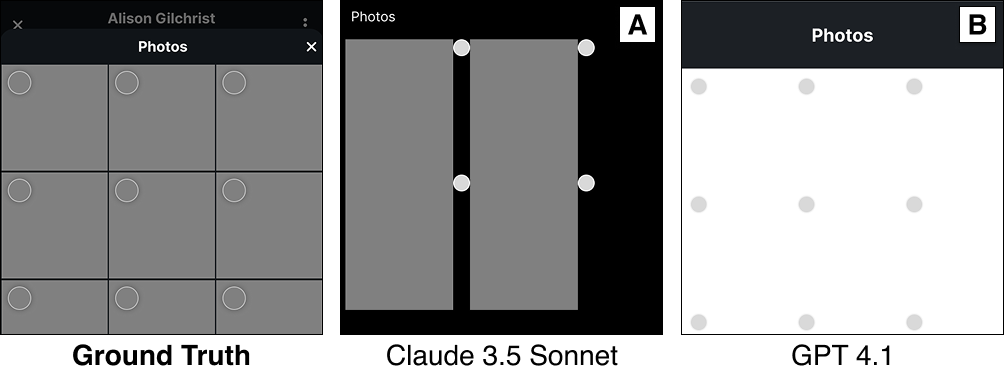}
  \caption{\textbf{Error case 2}. Updates to auto-layout settings trigger (A) disintegration between selector and placeholder or (B) push the placeholder downwards beyond the screen.}
  \label{fig:error-type-layout}
\end{figure}

Models reveal erroneous reasoning about screen auto‑layout. 
In Figma, this feature automatically realigns and resizes child elements when their parent resizes; similar mechanisms appear in other editors~\cite{figma2025auto, sketch2025stack}. 
Because any modification to the layout parameters can propagate through an entire hierarchy, the task demands prediction of changes without dictating their values through commands. 
As shown in Fig.~\ref{fig:error-type-layout}, the models' adjustment to auto-layout disintegrates the selector button and its placeholder (A) and forces the placeholder below the viewport (B).

\subsubsection{Text Operation Errors}

\begin{figure}[H]
  \centering
  \includegraphics[width=\columnwidth]{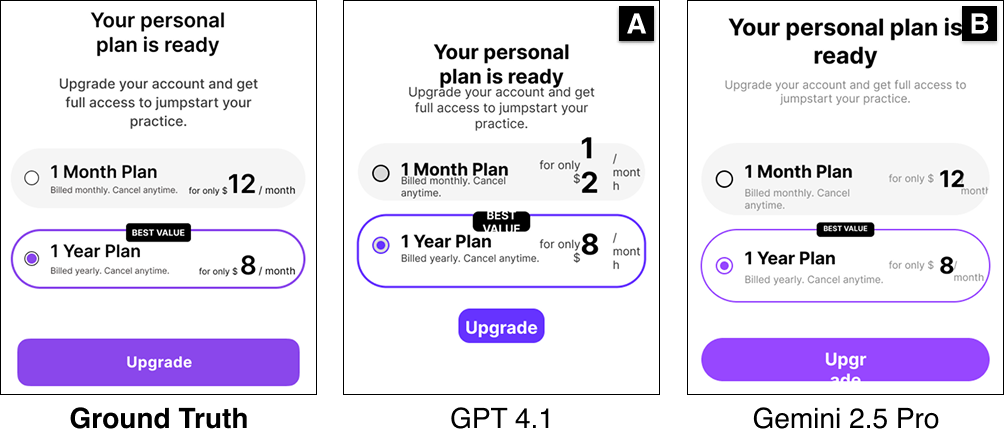}
  \caption{\textbf{Error case 3}. Models (A, B) fail to assign sufficient span to the text component and trigger text overflow.}
  \label{fig:error-type-text}
\end{figure}

Models commonly fail to infer appropriate dimensions from text attributes (e.g., font, size, and spacing), leading to narrowly sized text components and subsequent text overflow.
Figure~\ref{fig:error-type-text} illustrates two typical failures: (A) the price label extends beyond its bounding box, breaking visual alignment, and (B) the button label overruns its frame, blending into the background. 

\section{Conclusion}
\label{sec:conclusion}
We introduce \benchname{}, the first benchmark designed to evaluate vision-language models (VLMs) on tool-based UI design tasks. 
Our findings suggest that current state-of-the-art models exhibit promising capabilities in replicating and modifying interface designs through tool invocation with design software.
Particularly, high-performing models exhibit more diverse tool use and strategic behaviors.
Nonetheless, adverse tool selection and design errors highlight the limitations in the implementation framework and the model capacity.
By presenting an initial evaluation of VLMs performing design tasks within conventional software, \benchname{} offers valuable insights toward achieving human-aligned design automation with VLMs.

\section*{Acknowledgements}
\label{sec:acknowledgements}
We thank Chulwoo Kim, May Jorella Lazaro, and Ji Soo Yi from the CX Insight Team (MX Division), Samsung Electronics for insights on UI design workflows, and members of the KAIST Interaction Lab (KIXLAB) for their constructive feedback. 
This work was supported by the CX Insight Team (MX Division), Samsung Electronics, and by an IITP grant funded by the Korean government (MSIT) (No. RS-2024-00443251, Accurate, Safe, Multimodal, \& Multilingual Personalized AI Tutors).

\bibliography{aaai2026.bib}

\clearpage

\appendix
\section*{Appendix}
\label{sec:appendix}
\setcounter{secnumdepth}{2} 
\section{Data Collection Procedure}
To create a consistent dataset representative of the current design styles, we selected the Figma Community as our data source.
Given that Figma does not offer an officially documented API on the Figma Community, we devised a data collection procedure to collect and sample data relevant to the evaluation. 
Additionally, we implemented a task instruction generation step to provide instruction prompts for a model in the two design modification tasks: \emph{attribute update} and \emph{component insertion}, guiding the model to attain a narrowly scoped task goal.

\subsection{Data Selection}
We chose a manual selection process to identify Figma projects adherent to the criteria: each project was required to have at least two high-fidelity, editable UI designs. 
Each project is accessed through the ``Design resources'' tab under the category ``Mobile app design templates.'' 
Resources unrelated to UI design, such as prototyping kits, static images, device mockups, or design systems, were excluded.
The design selection terminated when further designs offered smaller variations relative to previously collected samples. 
Finally, the selected UI screens were exported through the Figma REST API, producing a set of SVG, PNG, and JSON data for each design.

\subsection{Data Sampling}
To analyze diversity within the dataset, we first categorized the dataset into 30 distinct UI types based on visual characteristics. 
The initial UI design categories were adapted from Mobbin’s web UI classifications. 
Due to the extensive nature of these categories, one of the authors used GPT-4.5 to generate initial clusters and manually refined the categories to minimize overlap and increase clarity. 
Finally, we used GPT-4.1-Mini with zero-shot prompting to automatically assign these categories to design images based on the predefined prompt.

\benchname{} addresses two primary tasks: UI design replication and modification, each requiring task-specific data selection methodologies. Modification tasks specifically demand manual selection based on clearly defined criteria. For example, tasks involving border radius require UI elements with rounded borders.

\begin{itemize}
    \item \textit{Design Replication}: We conducted stratified sampling based on the function-based UI type (e.g., login screen) to select 298 designs from the total of 3,327 categorized designs, ensuring balanced representation of different UI designs.
    This resulted in 10 examples per category, except for one category, limited to 8 examples due to insufficient data.
    \item \textit{Design Modification}: Authors manually selected 100 designs per modification type (attribute modification, component insertion, mode change) from the dataset. 
    Algorithmic selection was inadequate, as it could not reliably ensure the presence of necessary UI features (e.g., matching dark and light mode designs). 
    Authors manually generated target states of the modification by directly editing selected designs in Figma, such as removing a button.
\end{itemize}

\subsection{Data Annotation}

Each modification task includes a ground-truth design paired with a case-specific instruction. 
These instructions are essential for precisely guiding the model to select and modify specific design elements (e.g., adjusting the corner radius of an individual list element). 
Without explicit instructions, the model must infer the necessary edits by visually comparing the original and target states, introducing a significant perceptual bottleneck due to fine-grained visual differences. 
Specifically, instructions for \emph{attribute update} and \emph{component insertion} tasks are generated by comparing differences between the ``base'' and ``target'' states. 
GPT-4.1-Mini is employed for generating an instruction for each case. 
In contrast, instructions for the mode-change task are uniformly defined due to its general characteristic of applying color changes across multiple components.

\subsection{Model Collection Prompts}
\label{appendix:model-prompts}

\begin{figure}[h]
  \centering
  \caption{\textbf{Modification Task-1 Instruction Generation Prompt:} A prompt for generating task instructions for the modification Task-1 (Attribute Update). The keyword alternates based on the target components.}
  \begin{lstlisting}[
      basicstyle=\ttfamily\scriptsize,
      numberstyle=\tiny,
      frame=single,
      numbers=left,
      breakindent=0pt
    ]
** Instruction **
Write a human-like prompt instructing the model to transform the UI in the first image into the UI shown in the second image.
Specifically, focus on changing only the {colors|sizes|text contents|corner radius|positions} of UI elements from the first image to match those in the second image.
Then, explicitly identify and describe the required {color|size|text content|corner radius|position} changes for each UI element.
Start by requesting the modification on the image: "Make changes to the {colors|sizes|text contents|corner radius|positions} of the following UI elements:".
Then outline the changes in a structured format strictly using bullet points.

** Rules **
- Avoid specifying exact color codes, font names, dimensions, or other numerical values.
- You don't need to mention the image, just say "Make changes to the colors of the following UI elements:".
- You don't mention elements that are not changed.
  \end{lstlisting}
\end{figure}

\begin{figure}[h]
  \centering
  \caption{\textbf{Modification Task-2 Instruction Generation Prompt:} A prompt for generating a task instruction for the modification Task-2 (Component Insertion).}
  \begin{lstlisting}[
      basicstyle=\ttfamily\scriptsize,
      numberstyle=\tiny,
      frame=single,
      numbers=left,
      breakindent=0pt
    ]
** Instruction **
Write a human-like prompt instructing the model to add a component to the UI in the first image to make the UI shown in the second image.
Specifically, you are provided with a third image that illustrates the component to be added.
Comprehensively describe the component to be added using the third image.
If the third image contains text, include the text in the description.
Then suggest where to place the component in the first image to match the second image.
Start by requesting the insertion of the component. Start with "Insert the following component into the UI:"."
Then outline the component in a structured format, strictly using bullet points.
Next, describe the placement of the component in the first image. Start with "Place the component in the following location:"."
Then describe the location in a structured format, strictly using bullet points.

** Rules **
- Avoid specifying exact color codes, font names, dimensions, or other numerical values.
- Explicity mention any text in the component. Make sure you do not miss any text.
- You should not mention the image in the instruction. Just elaborate on how to add the component.
- You should not mention elements that are not changed.
  \end{lstlisting}
\end{figure}

\begin{figure}[h]
  \centering
  \caption{Data categorization prompt for User Interface (UI) type classification}
  \begin{lstlisting}[
      basicstyle=\ttfamily\scriptsize,
      numberstyle=\tiny,
      frame=single,
      numbers=left,
      breakindent=0pt
    ]
###  SYSTEM  ###
You are an expert product-designer-turned-taxonomist.  
Your task is to inspect one UI screenshot and decide which single category below best describes the *purpose of the entire screen* (not individual widgets).

Available categories  
(only one may be returned):

1. **Onboarding & Account Creation** -- Account Setup, Splash Screen, Welcome/Get-Started, Sign-up, Walk-through  
2. **Authentication & Access** -- Login, Forgot-Password, Verification, Delete/Deactivate Account  

(...categories 3-29 omitted for brevity)

30. **Others** -- Any screen that clearly doesn't fit categories 1-29

###  INSTRUCTIONS  ###
1. Look at the screenshot holistically: What is the user trying to do here?  
2. Pick the single *most appropriate* category (or **30** if none fit).
  \end{lstlisting}
\end{figure}

\section{Data Refinement Procedure}
\label{appendix:data-refinement-procedure}

To improve the reliability of our dataset, we performed systematic refinement steps.
The procedure aims to align the visual appearance with the underlying structure, ensuring that the design is reproducible via tool invocation, solely based on the image.
Initially, we identified significant noise in the dataset, manifesting as inconsistent patterns in the structure of common UI elements (e.g., a large gradient background hidden under button elements that extends over the parent frame).
This inconsistency poses a critical challenge for evaluation, as the structure of the ground truth design exhibits an arbitrary pattern.
Therefore, our refinement procedure focuses on matching the visual appearance to an underlying structure and ensuring the design is reproducible solely via tool invocation.
We expect the outlined refinement procedure to help future researchers plan similar data refinement procedure using design platforms and software.

\subsection{Data Analysis}

With the initial review of the dataset, we conducted automated data cleansing procedures. 
Specifically, (i) all original images, such as profile photos, were replaced with grey placeholders due to limitations in reproducing these images within the design software.
(ii) Locked design elements were unlocked to enable manipulation.
(iii) All component instances in Figma, which are copies of reusable components synchronized with their originals, were detached to allow independent editing.
(iv) Hidden elements, which are visually irrelevant, were removed entirely.
(v) Fonts were standardized uniformly to the ``Inter'' typeface while preserving original stylistic attributes, including weight and style. 
This was due to accessibility constraints associated with certain font families in different environments.

After automated cleansing, we performed a detailed analysis of the remaining issues that affect the model's capability to generate designs from input images successfully. 
Our evaluation identified three primary error categories that degrade data quality: (i) component occlusions, (ii) complex illustrations, and (iii) visual clutter.
Component occlusions describe cases where components are obscured by overlaying objects or positioned beyond the visible boundaries, such as modal dialogues with hidden background layers. 
Complex illustrations indicate highly detailed visuals that require a complex layering of multiple vectors, which falls outside the current user interface generation scope. 
Visual clutter refers to irregular layouts in which groups or components are slightly offset by small amounts (e.g., 2 pixels) while looking aligned. 
Additionally, our review revealed inaccuracies within the task modification instructions generated by GPT-4.1-mini.

\subsection{Manual Revision}

To enhance dataset reliability, we performed a manual review of each design involving four participants who (i) are currently majoring in a design or related field and (ii) have experience with the Figma tool. 
Participants were recruited through snowball sampling and compensated approximately USD 11 per hour, contributing a total of 7 hours each to the review process. 
Participants systematically checked for the three predefined issue categories and made necessary corrections. 
In modification tasks specifically, they validated the task instruction prompts and annotated the corresponding required tools. 
Additionally, each design was re-reviewed by the author, who applied further corrections when necessary. 
Out of 598 design tasks, our revision addressed component occlusions (200 cases, 33.4\%), complex illustrations (93 cases, 15.5\%), and visual clutter (174 cases, 29.0\%).

\subsubsection{Data Revision Instruction}
\label{appendix:design-refinment-instruction}

During the revision, we asked the four participants to conduct the manual revision.
Figure~\ref{fig:data-revision-prompt} shows the revision instruction that participants followed during the manual revision process.

\begin{figure}[h]
  \centering
  \caption{\textbf{Data Revision Instruction} An instruction given to expert designers for revising each task case.}
  \begin{lstlisting}[
      basicstyle=\ttfamily\scriptsize,
      numberstyle=\tiny,
      frame=single,
      numbers=left,
      breakindent=0pt
    ]
1. Remove Invisible Elements from the Frame (label: invisible)

    Removes elements that are hidden behind other objects or are otherwise not visible within the Figma frame.

    [Example Image]
    Example 1: The card element hidden beneath the keyboard needs to be removed.

2. Remove Complex Illustrations (label: illustration)

    Removes illustrations used for purely aesthetic purposes (e.g., pictures, drawings) as opposed to visual elements that convey information (e.g., icons).

    [Example Image]
    Example 1: The aesthetic illustration located in the center is removed.

3. Remove Element and Layer Clutters (label: clutter)

    Addresses design errors such as elements that are severely misaligned with others or extend beyond the boundaries of the Figma frame.

    [Example Image]
    Example 1: A case where individual elements are subtly shifted and not correctly arranged along the alignment axis.

    [Example Image]
    Example 2: A case where an element protrudes beyond its parent frame due to incorrect grouping.
  \end{lstlisting}
  \label{fig:data-revision-prompt}
\end{figure}

\subsubsection{Design Issue Examples}
\label{appendix:design-error-examples}
\begin{figure}[h]
  \centering
  \includegraphics[width=\columnwidth]{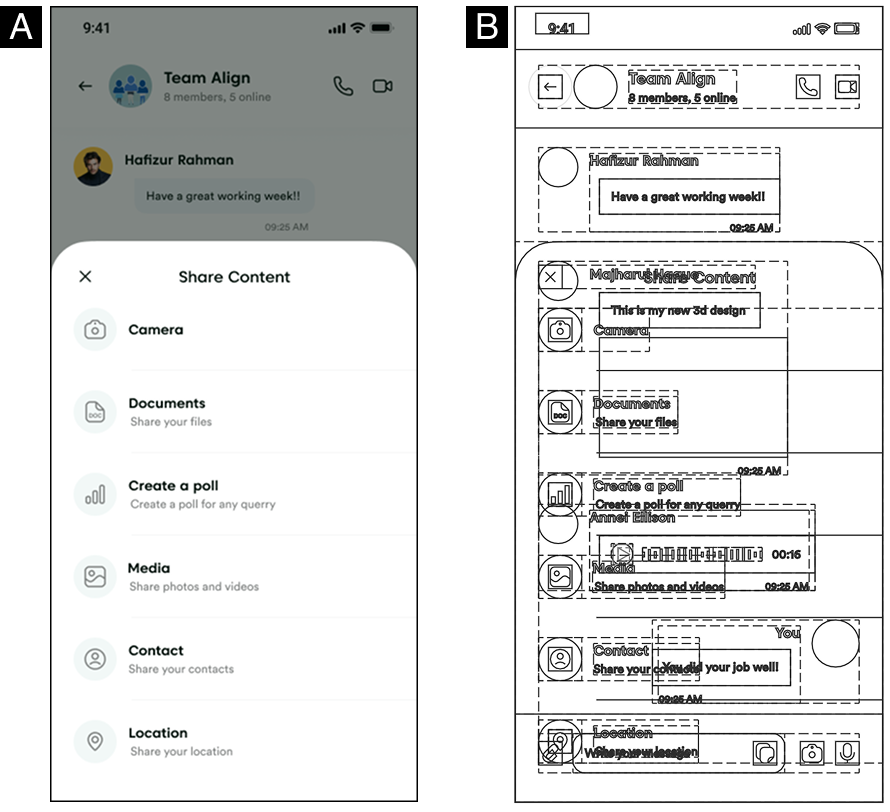}
  \caption{\textbf{Case: Component Occlusion}: (A) A UI design image showing a partial message list above a bottom sheet. (B) The underlying structure of the design reveals the components hidden beneath the bottom sheet.}
  \label{fig:design-error-invisible}
\end{figure}

\begin{figure}[h]
  \centering
  \includegraphics[width=\columnwidth]{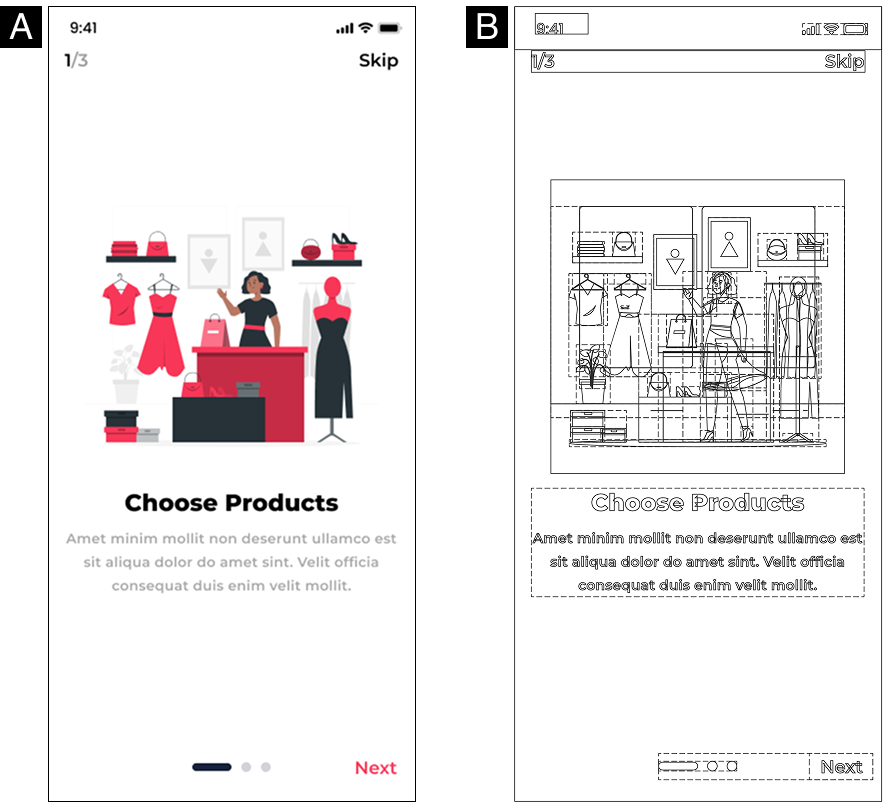}
  \caption{\textbf{Case: Complex Illustration}: (A) A UI design image with complex vector-based illustration at the center of the screen; (B) The underlying structure of the design reveals the complex layers and vector paths.}
  \label{fig:design-error-illustration}
\end{figure}

\begin{figure}[h]
  \centering
  \includegraphics[width=\columnwidth]{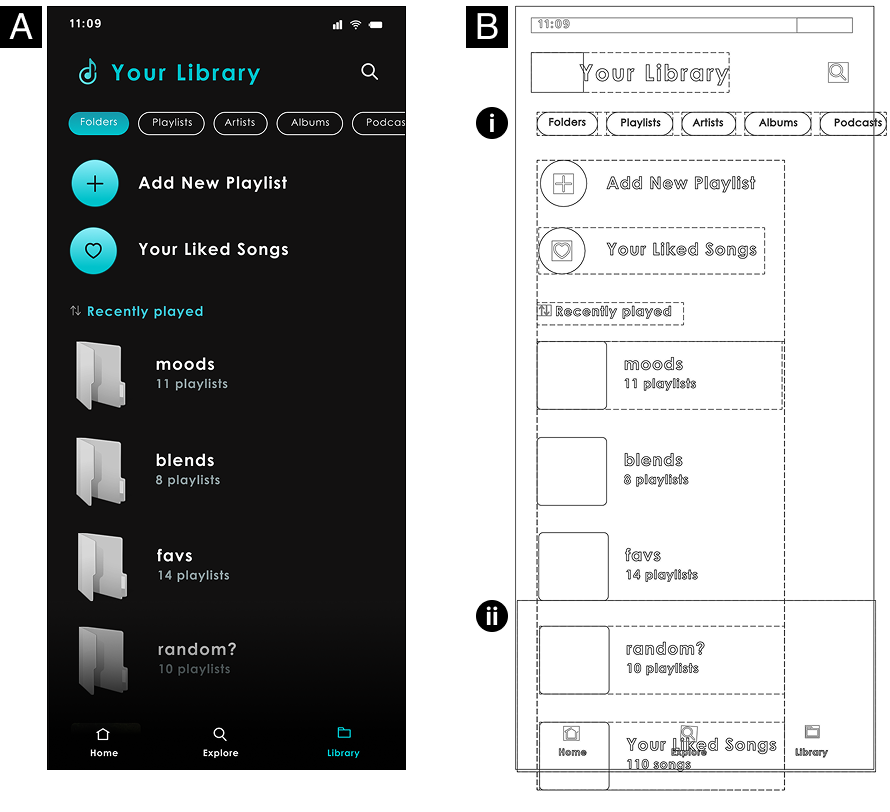}
  \caption{\textbf{Case: Visual Clutters}: (A) A UI design image with visually aligned components; (B) The underlying structure of the design shows (i) an element overflowing its container and (ii) child components that are misaligned with their parent by few pixels.}
  \label{fig:design-error-layer-clutters}
\end{figure}

We identify three major issues regarding the collected Figma design examples: component occlusions (Fig.~\ref{fig:design-error-invisible}), complex illustrations (Fig.~\ref{fig:design-error-illustration}), and visual clutter (Fig.~\ref{fig:design-error-layer-clutters}).

\section{Data Characteristics}
\label{appendix:data-statistics}

\subsection{Dataset Overview}

This appendix provides a detailed overview of the dataset compiled for our experiments. 
The dataset was collected from the Figma community, and each sample is composed of metadata, a JSON structure, a PNG image, and an SVG image. 
The metadata, summarized in the Table~\ref{tab:dataset-metadata} below, includes details such as the source URL, license, design type, and structural properties. 
All designs are distributed under the CC BY 4.0 license (see Fig.~\ref{tab:dataset-source} for data source).

\begin{table}[h]
  \centering
  \small
  \setlength{\tabcolsep}{4pt}
  \renewcommand{\arraystretch}{1.05}
  \caption{Common Metadata Attributes}
  \label{tab:dataset-metadata}

  \begin{tabularx}{\linewidth}{@{}l X@{}}
    \toprule
    \textbf{Field} & \textbf{Description} \\ \midrule
    url              & Link to the original Figma file or prototype. \\
    license          & Concise statement of the file’s usage license. \\
    node\_id         & Unique identifier of the parent frame in Figma. \\
    type\_id         & Numeric code denoting the design category. \\
    type\_name       & Human-readable name of that design category. \\
    node\_depth      & Maximum nesting depth in the structure JSON. \\
    node\_count      & Total number of nodes in the structure JSON. \\
    node\_type\_count & Per-type node counts in the JSON. \\
    \bottomrule
  \end{tabularx}
\end{table}

The data is organized into three distinct subsets for different evaluation tasks: (i) the Source Dataset, (ii) the Replication Task Set, and (iii) the Modification Task Set. 
The specific characteristics of each are detailed below.

\paragraph{Source Dataset}
The Source Dataset contains 3,327 original design examples parsed directly from projects in the Figma Community, encompassing 30 different UI design types.
Table~\ref{tab:uicategories} shows the distribution of the type in the Source Dataset.
We observed the gradual decline pattern in the frequency per UI type, with the largest group, ``Onboarding \& Account Creation,'' revealing 398 examples and ``Pricing \& Paywall'' showing 8 examples only.

\begin{table*}[h]
  \centering
  \small
  \setlength{\tabcolsep}{4pt}
  \renewcommand{\arraystretch}{1.05}
  \caption{UI Categories and Frequency for Source Dataset}
  \label{tab:uicategories}

  \begin{tabularx}{\textwidth}{@{\extracolsep{\fill}} l r X}
    \toprule
    \textbf{Category} & \textbf{Count} & \textbf{Exemplary components} \\ \midrule
    Onboarding \& Account Creation   & 398 & Account setup, splash screen, welcome/get-started, sign-up, walk-through \\
    Home \& Feeds                    & 397 & Home, news/social feed, browse/discover hub \\
    CRUD (Create–Update–Delete)      & 316 & Add/create, edit form, delete confirmation, draw/annotate \\
    Authentication \& Access         & 254 & Login, forgot-password, verification, delete/deactivate account \\
    Messaging \& Email               & 230 & Chat thread, inbox, bot conversation \\
    Permissions \& Security          & 211 & OS permission request, two-factor verification prompt, privacy policy \\
    Content Detail                   & 201 & Article/product/movie/note detail \\
    Profile \& Personal Info         & 183 & My account/profile, user/group profile, followers/following \\
    Feedback \& Success States       & 137 & Success confirmation, acknowledgement banner, “Done!” dialogs \\
    Cart \& Checkout                 & 128 & Cart/bag, checkout flow, order confirmation \\
    Search \& Explore                & 117 & Search results, suggestions, discover tab \\
    Orders \& Billing                & 103 & Order history, billing, wallet/balance, payment method \\
    Settings \& Preferences          & 85  & Global settings, dark-mode toggle, language selector \\
    Notifications \& Alerts          & 85  & Notification feed, in-app badge list, pull-to-refresh alert \\
    Loading \& Progress              & 62  & Spinner, skeleton loader, progress bar \\
    Dashboards \& Analytics          & 60  & Dashboard, charts, metric cards \\
    Select \& Input                  & 52  & Pickers (date, time, list…), set-value dialog \\
    Guided Tours \& Walk-throughs    & 32  & Coach marks, feature tutorial, step-by-step guide \\
    Community \& Groups              & 32  & Groups, invite teammates, leaderboard, achievements \\
    Promotions \& Rewards            & 32  & Coupons, rewards center, gamified rewards \\
    Filter \& Sort                   & 30  & Filter panel, sort dropdown, advanced filter sheet \\
    Error \& Empty States            & 30  & 404/error page, empty state, permission denied \\
    Social Engagement                & 24  & Like/up-vote, comments, share sheet, follow/subscribe \\
    Learning \& Tasks                & 20  & Quiz/lesson detail, stories, goal/task tracker \\
    Utilities \& Productivity        & 17  & Calendar, reminder, timer, schedule manager \\
    Others                           & 17  & Components that do not fit the above groups \\
    Media Playback                   & 36  & Audio/video player, now-playing screen \\
    Maps \& Location                 & 15  & Map view, address form, nearby results \\
    Media Capture \& Editing         & 15  & Camera/scanner, media editor, recorder, filters \\
    Pricing \& Paywall               & 8   & Pricing table, paywall, promotions/rewards upsell \\
    \bottomrule
  \end{tabularx}
\end{table*}

\paragraph{Replication Task Set}
The Replication Task Set is designed to evaluate the model's ability to reproduce existing designs. 
It contains 298 cases created by applying stratified sampling to the Source Dataset. We sampled 10 examples for each of the 30 UI types, except for the ``Pricing \& Paywall category,'' where all 8 available source examples were used. 
Each sample in this set includes an additional metadata field, ``difficulty,'' which is classified as either standard or hard.

\paragraph{Modification Task Set}
The Modification Task Set is designed to assess the model's capacity for targeted design editing. 
It consists of 300 cases, with 10 examples for each of the 30 UI types. 
The metadata for each case includes an instruction field that provides a task-specific description of the required change. 
The data provided per case varies by task type:

\begin{itemize}
    \item For attribute update and mode change tasks, each case includes a \texttt{base\_state} and a \texttt{target\_state}.
    \item For component insertion tasks, each case includes a \texttt{base\_state}, the \texttt{component} to be inserted, and the \texttt{target\_state}.
\end{itemize}

\subsection{Node Structure}

Each design case is represented by a JSON file that encodes its hierarchical node structure. 
This representation is derived from the Figma REST API, which provides access to the node structure in a specific design, with minor modifications to ensure consistency with the parser based on the Figma Plugin API. 
The JSON mirrors the layered nature of the design process; designers often construct interfaces by stacking, nesting, and grouping components.
Consequently, the data follows a tree-like structure where child nodes are recursively accessed through a children list.

Although this hierarchical structure resembles markup languages like HTML or SVG, its semantic purpose is fundamentally different. 
While markup is optimized for browser rendering, this JSON representation is intended to capture the design's structure as a designer perceives it. 
This distinction is evident in the prevalence of design-specific properties such as multiple fills and blend modes, which appear more frequently and prominently here than in typical web-oriented markup.

\subsubsection{Node as a Complexity Measure}

Given that each node corresponds to a design component, the total node count serves as a practical proxy for measuring design complexity. 
Higher node count indicates greater structural complexity, which primarily equates to greater visual complexity. 

\begin{figure}[h]
  \centering
  \includegraphics[width=\columnwidth]{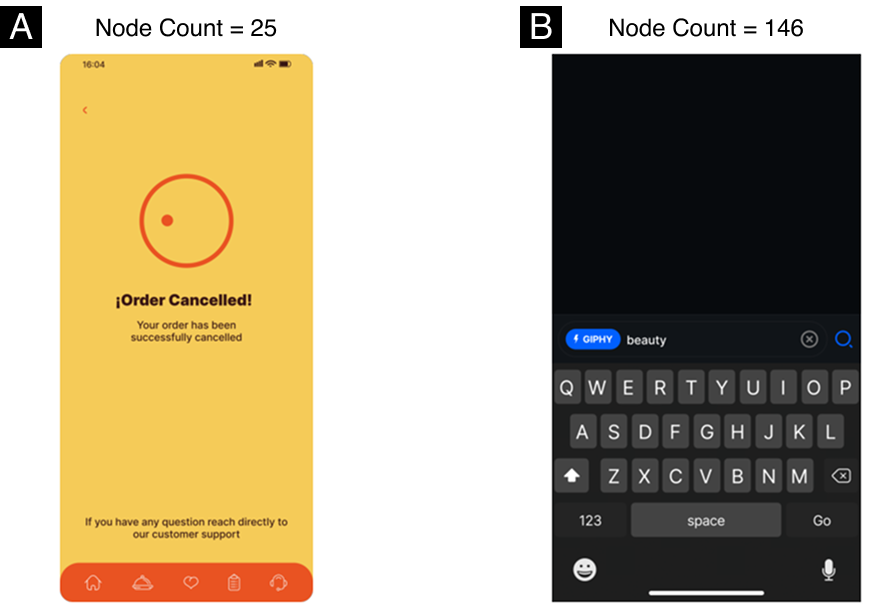}
  \caption{\textbf{Node Count Examples} Two representative designs randomly sampled from the (A) bottom 15\% and (B) top 15\% of node counts, respectively.}
  \label{fig:design-complexity-example}
\end{figure}

For instance, Figure~\ref{fig:design-complexity-example} contrasts two designs from the bottom and top 15th percentiles of node counts in our dataset. 
Design (A), with a low node count ($n=25$), features a simple layout with a few primary components, such as icons in a top bar and buttons in a bottom bar. 
Design (B) has a significantly higher node count ($n=148$), largely due to numerous fine-grained components, such as the individual keys of a virtual keyboard, which illustrates its greater structural complexity.

\subsection{Structural Analysis of Datasets}

As shown in Table~\ref{tab:per-dataset-structure}, the Replication Task Set and the Modification Task Set exhibit different patterns compared to the Source Dataset.
The Source Dataset is considerably more complex than these task sets. 
Its mean node count is approximately 65\% and 83\% greater than that of the replication and modification sets, respectively. 
Furthermore, its designs feature a deeper average hierarchy, with a mean node depth of 7.06±2.32, compared to approximately 6.5 for the other two sets.
The numbers reveal a progressive simplification of designs from the originally collected to the more refined examples in the replication and modification test sets.

\begin{table}
\centering
\caption{
  Structural Statistics Across Datasets
}
\label{tab:per-dataset-structure}
\begin{adjustbox}{max width=\columnwidth}
\begin{tabular}{
  l 
  >{\centering\arraybackslash}p{0.26\columnwidth}
  >{\centering\arraybackslash}p{0.26\columnwidth}
  >{\centering\arraybackslash}p{0.26\columnwidth}
}
\toprule
\textbf{Attribute} & \textbf{Original} & \textbf{Replication} & \textbf{Modification} \\
\midrule
Node Depth & 7.06 (2.32) & 6.48 (2.04) & 6.62 (2.40) \\
Node Count & 126.25 (194.30) & 76.30 (46.81) & 69.11 (49.40) \\
\bottomrule
\end{tabular}
\end{adjustbox}

\begin{tablenotes}[flushleft]
\footnotesize
\item Each cell reports the mean and standard deviation (SD).
\end{tablenotes}

\end{table}


Furthermore, we analyze the complexity of the designs in our datasets by examining the distributions of node count, depth, and type. 
This analysis helps to characterize the impact of our data revision process.

\begin{figure}[h]
  \centering
  \includegraphics[width=\columnwidth]{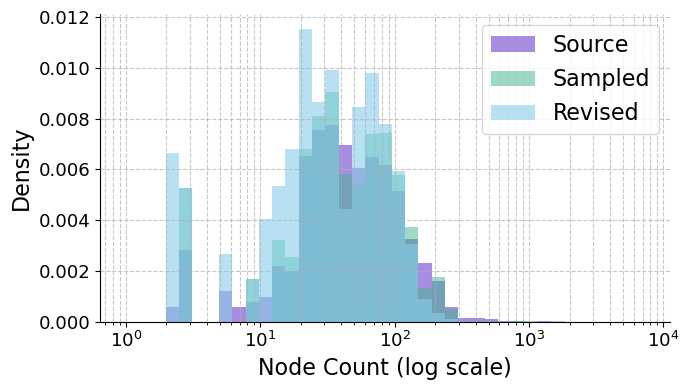}
  \caption{\textbf{Node Count Distribution} A distribution of the node count for the Source Dataset and Replication Task Set (two states: sampled and revised).}
  \label{fig:source-dataset-node-count-distribution}
\end{figure}

Figure~\ref{fig:source-dataset-node-count-distribution} compares the distribution of node counts for the Source Dataset alongside the Replication Task Set before (sampled) and after (revised) our data revision process. 
Initially, the distribution of the sampled dataset closely mirrors that of the source dataset, confirming the effectiveness of our stratified sampling strategy. 
However, after the revision, the distribution for the revised data exhibits a noticeable leftward shift. 
This shift indicates that our revision process successfully reduced the overall complexity of the designs by decreasing the number of nodes per design.

\begin{figure}[h]
  \centering
  \includegraphics[width=\columnwidth]{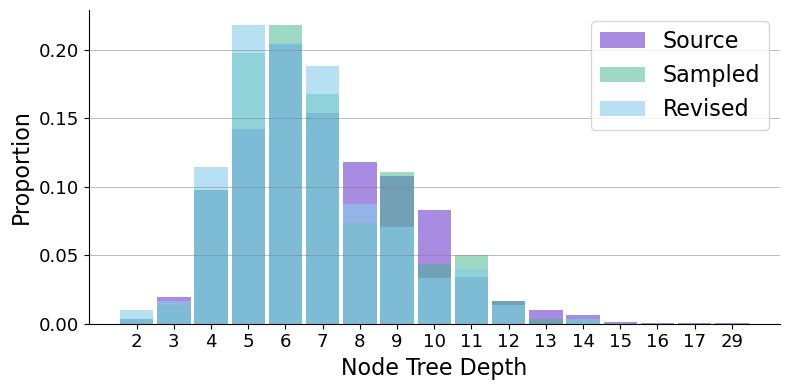}
  \caption{\textbf{Node Depth Distribution} A distribution of the node depth for the Source Dataset and Replication Task Set (two states: sampled and revised).}
  \label{fig:source-dataset-node-depth-distribution}
\end{figure}

In terms of node depth, the distributions for all three sets (source, sampled, and revised) are largely similar, as shown in Figure~\ref{fig:source-dataset-node-depth-distribution}. 
All distributions are centered around a depth of 5 to 7 and resemble a Gaussian distribution. 
Unlike the node count, the revision process appears to have had a less pronounced effect on node depth. 
This suggests that the complexity reduction was achieved by simplifying the node structure rather than by reducing their maximum depth.

\begin{figure}[h]
  \centering
  \includegraphics[width=\columnwidth]{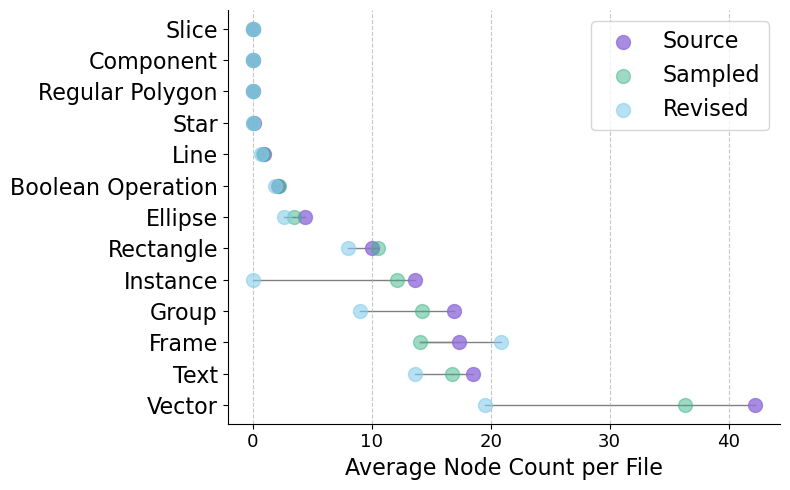}
  \caption{\textbf{Node Type Distribution} A distribution of the node type for the Source Dataset and Replication Task Set (two states: sampled and revised).}
  \label{fig:source-dataset-node-type-distribution}
\end{figure}

Figure~\ref{fig:source-dataset-node-type-distribution} details the average count of different node types across the datasets. 
The source and sampled datasets exhibit similar node type distributions. In contrast, the revised dataset displays a distinct pattern resulting from our revision protocol. 
Specifically, the counts of \texttt{Instance} and \texttt{Vector} nodes are significantly reduced. This directly corresponds to our targeted revision goals: to detach instance components for direct manipulation and to simplify complex vector graphics.
Concurrently, we observe an increase in \texttt{Frame} nodes and a decrease in \texttt{Group} nodes, reflecting the detached instances (which become a frame) and the removal of the visual clutter.

\subsection{Node Structure Example}
\label{appendix:figma-node-structure-example}

Figure~\ref{fig:json-structure-example} illustrates the JSON format used to represent the design structure. 
The data is organized hierarchically, where each node is nested as a child of its parent. 
Every node contains a unique ID and a type attribute. 
Visual characteristics are specified through properties such as \texttt{blendMode} and \texttt{fills}, which define the node's appearance.

\begin{figure}[h]
  \centering
  \caption{\textbf{Design Structure JSON Example} An example formatting for design structure JSON.}
  \begin{lstlisting}[
      basicstyle=\ttfamily\scriptsize,
      numberstyle=\tiny,
      frame=single,
      numbers=left,
      breakindent=0pt
    ]
"children": [
  {"id": "I1:1383;11:211",
    "name": "Image",
    "type": "INSTANCE",
    "children": [
      {"id": "I1:1383;11:211;11:209",
        "name": "Shape",
        "type": "VECTOR",
        "blendMode": "PASS_THROUGH",
        "fills": [
          {"blendMode": "NORMAL",
            "type": "SOLID",
            "color": {
              "r": 0.7041666507720947,
              "g": 0.8579999804496765,
              "b": 1.0,
  \end{lstlisting}
  \label{fig:json-structure-example}
\end{figure}

\clearpage

\onecolumn 
{\small
\begin{longtable}{p{0.32\textwidth}p{0.18\textwidth}p{0.36\textwidth}p{0.10\textwidth}}
    \caption{Figma Community UI Design Data Sources\label{tab:dataset-source}}\\
    \textbf{Title} & \textbf{Author} & \textbf{URL} & \textbf{Last Access} \\
    \hline
    \endfirsthead
    \textbf{Title} & \textbf{Author} & \textbf{URL} & \textbf{Last Access} \\
    \hline
    \endhead
    Mobile Apps – Prototyping Kit & Renata Pôrto & \url{https://www.figma.com/community/file/1129468881607079432} & 2025-08-03 \\
    WhatsApp UI Screens & Pixsellz & \url{https://www.figma.com/community/file/874576344344319149} & 2025-08-03 \\
    Food Delivery App & Yamesh\_SB & \url{https://www.figma.com/community/file/1231521889522325040} & 2025-08-03 \\
    Meditation app UI & Afsar & \url{https://www.figma.com/community/file/882888114457713282} & 2025-08-03 \\
    Online Groceries App UI & Afsar & \url{https://www.figma.com/community/file/882645007956337261} & 2025-08-03 \\
    Open Fashion - Free eCommerce UI Kit & UI Store Design & \url{https://www.figma.com/community/file/1055151140671808467} & 2025-08-03 \\
    eCommerce App UI Kit - Case Study Ecommerce Mobile App UI kit & UI-UX Expert (Aashifa) & \url{https://www.figma.com/community/file/1264098337558102933} & 2025-08-03 \\
    Schedule Management Platform (2019) & Ilia Utkin & \url{https://www.figma.com/community/file/904740192549800638} & 2025-08-03 \\
    Dokterian - Doctor Appointment Mobile App & Lokanaka Studio & \url{https://www.figma.com/community/file/1106038596434269509} & 2025-08-03 \\
    Telegram UI Screens & Pixsellz & \url{https://www.figma.com/community/file/874574404452104362} & 2025-08-03 \\
    iBank - Banking \& E-Money Management App | FinPay | Digital | Finance Mobile Banking App Ui Kit & Seju & \url{https://www.figma.com/community/file/1322236579213422290} & 2025-08-03 \\
    Mobile Templates & Gareth Lund & \url{https://www.figma.com/community/file/1076609425686201098} & 2025-08-03 \\
    Shoppe - eCommerce Clothing Fashion Store Multi Purpose UI Mobile App Design & Joy Patel & \url{https://www.figma.com/community/file/1321464360558173342} & 2025-08-03 \\
    Twitter UI Screens & Pixsellz & \url{https://www.figma.com/community/file/874600772514053297} & 2025-08-03 \\
    20 Screen Login \& Register Mobile App & Ali Husni Majid & \url{https://www.figma.com/community/file/1370757927948360864} & 2025-08-03 \\
    Facebook Messenger UI Screens & Pixsellz & \url{https://www.figma.com/community/file/874577850804632750} & 2025-08-03 \\
    Invest App App design & Marvis & \url{https://www.figma.com/community/file/908105441861974884} & 2025-08-03 \\
    Hoteliq - Booking Hotel App Design & Bony Fasius Gultom & \url{https://www.figma.com/community/file/1169928945460966636} & 2025-08-03 \\
    Fitness App UI Kit for Gym Workout App Fitness Tracker Mobile App Gym Fitness Mobile App UI Kit & Vectorfair & \url{https://www.figma.com/community/file/1356281690251535631} & 2025-08-03 \\
    BrainBox Ai ChatBot Mobile App Full 100\% Free UI Kit & Divus Motion & \url{https://www.figma.com/community/file/1316069776447695879} & 2025-08-03 \\
    Food Delivery App UI with Illustrations & Blush & \url{https://www.figma.com/community/file/989103752998044165} & 2025-08-03 \\
    Miquido - Gym Chatbot App Concept & Miquido & \url{https://www.figma.com/community/file/1124291019178940578} & 2025-08-03 \\
    Online Learning App Design & Benzatine Infotech & \url{https://www.figma.com/community/file/1134363994710610029} & 2025-08-03 \\
    Expenio - Personal Finance UI Kit & Priyank & \url{https://www.figma.com/community/file/885503108183774962} & 2025-08-03 \\
    Calendar mobile app & Itai Bracha & \url{https://www.figma.com/community/file/1087092786661212228} & 2025-08-03 \\
    Food App Design UI Template & DS CODE & \url{https://www.figma.com/community/file/1362393407429980800} & 2025-08-03 \\
    Mobile E-Learning App Design & Bony Fasius Gultom & \url{https://www.figma.com/community/file/1116625179283253250} & 2025-08-03 \\
    Musium - Music App UI & Chandrama Saha & \url{https://www.figma.com/community/file/1143115506742537849} & 2025-08-03 \\
    Food Delivery App UI Kit Food App Design Food Mobile App Delivery UI & Vectorfair & \url{https://www.figma.com/community/file/1350568717710815378} & 2025-08-03 \\
    Medical Health Mobile App Dermatology App Ui Kit Doctor Mobile App & Vectorfair & \url{https://www.figma.com/community/file/1343379534839915650} & 2025-08-03 \\
    Messaging - Chatbox App Design & Benzatine Infotech & \url{https://www.figma.com/community/file/1171349248796492917} & 2025-08-03 \\
    Fitness App & Nickelfox Design & \url{https://www.figma.com/community/file/1134379406586739829} & 2025-08-03 \\
    Finance Management Mobile App UI UX Kit for Budget Tracker Financial Prototype Design & Vectorfair & \url{https://www.figma.com/community/file/1358180178353418071} & 2025-08-03 \\
    Transferme Banking Financial Full APP Ui Template Free 57 plus screen & Divyesh Patel & \url{https://www.figma.com/community/file/1304389317046750323} & 2025-08-03 \\
    Meal Planner App UI Kit & Jordi & \url{https://www.figma.com/community/file/1042726207857424115} & 2025-08-03 \\
    Ai Food Delivery- App Full UI Kit Free 280+ Screen & Divyesh Patel & \url{https://www.figma.com/community/file/1313930988976055629} & 2025-08-03 \\
    Mental Health Fitness Mobile App design & Nickelfox Design & \url{https://www.figma.com/community/file/1029623608002086997} & 2025-08-03 \\
    Crypto Trading App UI Kit & Agilan KS & \url{https://www.figma.com/community/file/987218729121549341} & 2025-08-03 \\
    Task manager Mobile app Ui & Rohan Kumar & \url{https://www.figma.com/community/file/1025672621370192050} & 2025-08-03 \\
    Travel APP & Al Razi Siam & \url{https://www.figma.com/community/file/1167020373713392658} & 2025-08-03 \\
    Travenor - Travelling App & Atif Nadeem & \url{https://www.figma.com/community/file/1323887283985127984} & 2025-08-03 \\
    Budgeting mobile app design & Khoa (JAK) & \url{https://www.figma.com/community/file/1068075442755755994} & 2025-08-03 \\
    Timon - Senior Booking App & Atif Nadeem & \url{https://www.figma.com/community/file/1168885651980114323} & 2025-08-03 \\
    Mobile App Wireframing UI Kit & Vlad Solomakha & \url{https://www.figma.com/community/file/1344366595037131653} & 2025-08-03 \\
    Grocery App (Big Cart) & Usama Memon & \url{https://www.figma.com/community/file/1180139584868792437} & 2025-08-03 \\
    Travelin Travelling App Ui & Dha Adhi & \url{https://www.figma.com/community/file/1377526482587017983} & 2025-08-03 \\
    TikTok UI 2024 & Piero Borgo & \url{https://www.figma.com/community/file/1181613055862447288} & 2025-08-03 \\
    Health Fitness Workout App (FREEBIE - Prototype) & Sajid & \url{https://www.figma.com/community/file/977207453380268315} & 2025-08-03 \\
    Mobile Chat Kit & Francesco Mastrogiacomo & \url{https://www.figma.com/community/file/959248924999598689} & 2025-08-03 \\
    Linkedin UI Screens & Mikhail\_\_Kulinich & \url{https://www.figma.com/community/file/1076183068526016019} & 2025-08-03 \\
    Fitness Mobile App UI & Arash RQ & \url{https://www.figma.com/community/file/1123006530284948493} & 2025-08-03 \\
    Grocery Shopping App & Yamesh\_SB & \url{https://www.figma.com/community/file/1235854850814701017} & 2025-08-03 \\
    Si - Sehat - Mobile UI Kit & dwiky setiawan & \url{https://www.figma.com/community/file/1341489112333878981} & 2025-08-03 \\
    WhatsApp Screens 2025 with Meta AI - UI for iOS & Simone Giannangeli & \url{https://www.figma.com/community/file/1371557934947420911} & 2025-08-03 \\
    Doctor Appointment App UI Kit & PigFig & \url{https://www.figma.com/community/file/1302566707232887096} & 2025-08-03 \\
    Medical Clinic Booking (Doctor Appointment) App UI Concept & Hamza Tariq & \url{https://www.figma.com/community/file/1295662439127422938} & 2025-08-03 \\
    Uber Redesign & Ivan Fadila & \url{https://www.figma.com/community/file/996237331421768142} & 2025-08-03 \\
    Ecommerce App & Ramiro Vasquez & \url{https://www.figma.com/community/file/1091083465902099133} & 2025-08-03 \\
    Food Couriers - Food Delivery App - UI/UX Design Case Study & El Kamcy Speaks & \url{https://www.figma.com/community/file/1205134361239673955} & 2025-08-03 \\
    Plant Care App & Nickelfox Design & \url{https://www.figma.com/community/file/1043052073261558633} & 2025-08-03 \\
    Simple Music Player & Sumit Singh & \url{https://www.figma.com/community/file/976054929108031412} & 2025-08-03 \\
    Task manager - Mobile App & Saber Ali & \url{https://www.figma.com/community/file/983247483237345112} & 2025-08-03 \\
    Medical Mobile App & Nickelfox Design & \url{https://www.figma.com/community/file/1172153496393189176} & 2025-08-03 \\
    Online Game Streaming Mobile App & Nickelfox Design & \url{https://www.figma.com/community/file/1134819478872959489} & 2025-08-03 \\
    Chatting App UI Kit Design | E-Chat | Figma & Vuong Huu Thien & \url{https://www.figma.com/community/file/1370509894547995600} & 2025-08-03 \\
    Bio.links - Mobile App & Bruno Rodrigues & \url{https://www.figma.com/community/file/872143726192128099} & 2025-08-03 \\
    FluxStore & InspireUI & \url{https://www.figma.com/community/file/1271666050130425220} & 2025-08-03 \\
    Food Delivery App UI Kit Design | Speedy Chow | Figma & Vuong Huu Thien & \url{https://www.figma.com/community/file/1364368331565295794} & 2025-08-03 \\
    E-Learning App UI Template & DS CODE & \url{https://www.figma.com/community/file/1351560567994429197} & 2025-08-03 \\
    Ecommerce App UI Kit (Freebie) & Hamza Naeem & \url{https://www.figma.com/community/file/1362309395455453748} & 2025-08-03 \\
    Employee Management Mobile App & ghanshyam & \url{https://www.figma.com/community/file/1220440698169252596} & 2025-08-03 \\
    Home Decor App Mobile UI Kit | Interior Design Decoration Mobile App & Vectorfair & \url{https://www.figma.com/community/file/1356748341249072709} & 2025-08-03 \\
    Fintech Mobile App Update1.0 & Hassan Ali & \url{https://www.figma.com/community/file/1283725118401003453} & 2025-08-03 \\
    UI HR Management \& Task Management for Mobile Apps & Haidar & \url{https://www.figma.com/community/file/1437028069454890066} & 2025-08-03 \\
    The Ordinary-Skincare Mobile App & Nickelfox Design & \url{https://www.figma.com/community/file/1104389391440619014} & 2025-08-03 \\
    Advanced Prototyping Add to cart for a coffee app & Olorunfemi John & \url{https://www.figma.com/community/file/1263730789675045212} & 2025-08-03 \\
    Foxcrypto - Crypto App & Nickelfox Design & \url{https://www.figma.com/community/file/1147402245634536123} & 2025-08-03 \\
    Recipe App Ui Kit Food Mobile Recipe Cooking App & Vectorfair & \url{https://www.figma.com/community/file/1370589699536863229} & 2025-08-03 \\
    Dummy Login \& Register -  iOS App (Free) & Irvan Wibowo & \url{https://www.figma.com/community/file/1038653754860715312} & 2025-08-03 \\
    Wisecrypto - Cryptocurrency App & Hub Academy & \url{https://www.figma.com/community/file/1048450455946221363} & 2025-08-03 \\
    BahasaKu E-learning local language ui kit & Angga Cahya & \url{https://www.figma.com/community/file/1082985450689546470} & 2025-08-03 \\
    Exergize - Fitness App & Nickelfox Design & \url{https://www.figma.com/community/file/1121711277404210349} & 2025-08-03 \\
    AR Museum Guide & Nickelfox Design & \url{https://www.figma.com/community/file/1043459494713403436} & 2025-08-03 \\
    lyft - Taxi Booking App & Yamesh\_SB & \url{https://www.figma.com/community/file/1232616996666773350} & 2025-08-03 \\
    Planta - E-Commerce app for Plant Enthusiast & Tomey Tran & \url{https://www.figma.com/community/file/1057288219093151235} & 2025-08-03 \\
    Cupid Arrow Dating App: iOS/Android UI/UX & Shoeb Kamal & \url{https://www.figma.com/community/file/1323701241602792905} & 2025-08-03 \\
    Shop - Ecommerce Mobile App & DjectStudio & \url{https://www.figma.com/community/file/1311717403915877513} & 2025-08-03 \\
    Grocery Mobile App UI Design \& Prototype & Rafat & \url{https://www.figma.com/community/file/1103368288888925886} & 2025-08-03 \\
    Chatbot AI & Aliasghar - Davoodi & \url{https://www.figma.com/community/file/1339143564730755965} & 2025-08-03 \\
    Cinema tickets booking app & Yuriy Apretov & \url{https://www.figma.com/community/file/1130751076587602192} & 2025-08-03 \\
    Bazar - Books Mobile App & DjectStudio & \url{https://www.figma.com/community/file/1314317856766658266} & 2025-08-03 \\
    Log in, Registration, Onboarding / Sign in \& Sign up / Forgot password & Viktoriia & \url{https://www.figma.com/community/file/1358076679903181574} & 2025-08-03 \\
    Educational App - Freebie & Massimiliano Bolognesi & \url{https://www.figma.com/community/file/920981176050856532} & 2025-08-03 \\
    Medico - Doctors Appointment \&  Consultation App UI Kits & Säeef & \url{https://www.figma.com/community/file/1367424436471392722} & 2025-08-03 \\
    FoodBreak - AI Food Delivery App Kit Demo & Mafalda Matias & \url{https://www.figma.com/community/file/1246040803607605711} & 2025-08-03 \\
    Loyalty Kit - Free & Bdyhm & \url{https://www.figma.com/community/file/1322561827767522207} & 2025-08-03 \\
    Chating PWA & Aizazullah Kalhoro & \url{https://www.figma.com/community/file/1198752417967297161} & 2025-08-03 \\
    Parcel Delivery App UI Kit & Chuonraksa & \url{https://www.figma.com/community/file/1328277989233272800} & 2025-08-03 \\
    Workout app UI Kits & Spark8 Studio & \url{https://www.figma.com/community/file/1306588084283181346} & 2025-08-03 \\
    Cook - Food \& Drink Delivery Mobile App UI Kit Free (Figma Community) & Jos Pham & \url{https://www.figma.com/community/file/1255795628849357440} & 2025-08-03 \\
    PetNest (Pet App) & Akansha Sharma & \url{https://www.figma.com/community/file/1071859412883065265} & 2025-08-03 \\
\end{longtable}
}
\twocolumn

\clearpage

\section{Benchmark Execution Setup}

\subsection{Experiment Configuration}

\begin{figure*}[h]
  \centering
\includegraphics[width=\textwidth]{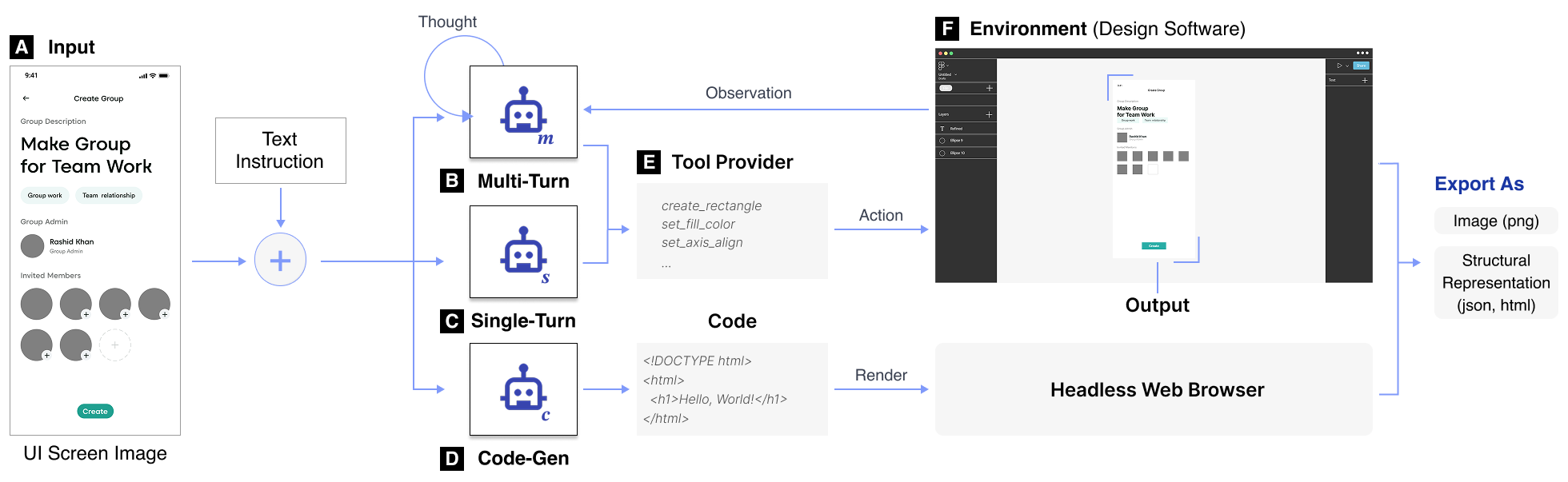} 
  \caption{\textbf{\benchname{} Experiment Pipeline}: (A) The pipeline feeds the model a target UI image with a textual task instruction. (B) Based on the input, the model generates a design via multi-turn thought–action–observation cycles with tool invocation; (C) single-turn tool invocation; or (D) direct UI code generation. For the code generation, we use Puppeteer to render the code as an image. (E) For the tool-based generation, the pipeline provides a 50-tool list for tool-based configuration. (F) Each tool is linked to a specific operation in Figma.}  
  \label{fig:appendix-pipeline}
\end{figure*}

We implement our experiment in three setups: tool-based multi-turn generation, tool-based single-turn generation, and code-based single-turn generation.
Figure~\ref{fig:appendix-pipeline} shows the pathways that each configuration produces a UI design image and the structural representation.
Our main experimental method, which constitutes the default \benchname{} configuration, is tool-based multi-turn generation. 
For our ablation study, we evaluate two additional conditions: tool-based single-turn generation and code-based single-turn generation. 
This uniform framework ensures that the results from all three methods are directly comparable.

\subsection{Execution Infrastructure}
\subsubsection{Pipeline Design}

\benchname{} adopts the Model-Context-Protocol (MCP) communication structure from an open-source project~\footnote{cursor-talk-to-figma-mcp, MIT license, \url{https://github.com/grab/cursor-talk-to-figma-mcp}}, enabling multi-turn tool invocation in Figma.
While inspired by its architecture, we redesigned the pipeline for research use. 
Instead of using the Cursor IDE as the MCP client, we implemented a custom client for controlled evaluation and agent orchestration. 
We also extended the tool interface, such as vector operations with explicit parameter control (e.g., shape, position, and color). 
These enhancements support interpretable, fine-grained manipulation of the environment, which is essential for benchmarking agent behavior in iterative UI design. 
Our structure enables both execution and structured analysis of reasoning and action quality in tool-driven settings.

Specifically, we implemented the communication between the model and the Figma environment using four key components: a Figma Plugin, a Socket Server, an MCP Server, and an MCP Client that interfaces with the model. 
A model interaction is triggered by a tool invocation, which follows this sequence:
\begin{itemize}
    \item The model's tool call is passed from the MCP Client to the MCP Server.
    \item The command is forwarded through the Socket Server to the Figma Plugin.
    \item The plugin executes the action on the Figma canvas.
    \item The action's outcome is returned to the model via the same path.
\end{itemize}

\subsubsection{ReAct Agent Implementation}

\begin{figure*}[h]
  \centering
\includegraphics[width=\textwidth]{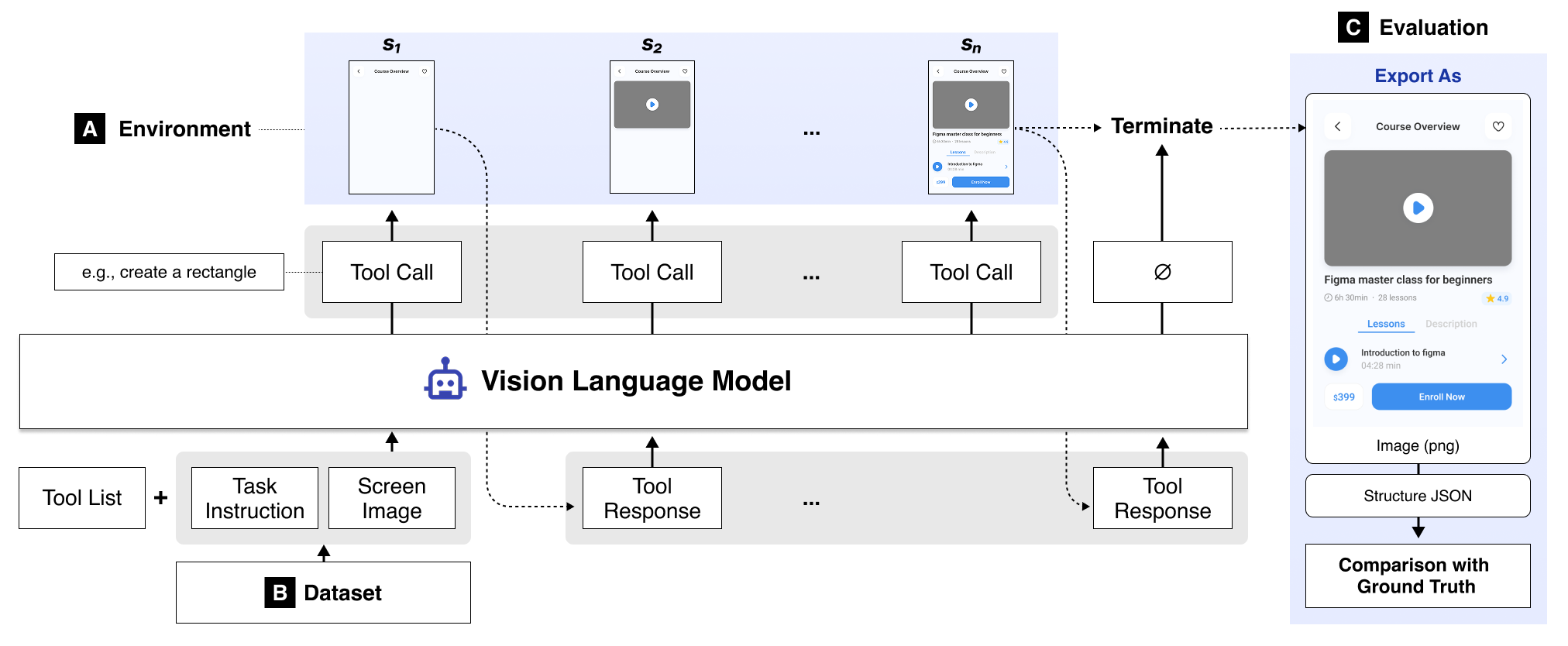} 
  \caption{\textbf{Detailed View of the ReAct Agentic Loop}: (A) A Vision-Language Model (VLM), acting as an agent, iteratively interacts with the design environment by invoking tools with the required arguments. This agentic loop continues until the VLM signals task completion by generating a final response that contains no further tool calls. (B) The initial input for each task is drawn from either the Replication or Modification Task Sets and consists of a natural language instruction and a corresponding ground-truth design image. (C) Upon task completion, \benchname{} evaluates the final generated design. The evaluation is performed by comparing both its rendered image and its underlying structural representation against the provided ground truth.}  
  \label{fig:appendix-agent-overview}
\end{figure*}

To enable iterative and multi-step design tasks, we employ an agentic process based on the ReAct framework. 
Figure~\ref{fig:appendix-agent-overview} shows the overview of the implementation.
The agent initiates an action (a tool invocation) by seeing the task instruction and the ground truth image.
Following each action, the model receives an observation (a response) from the Figma environment. 
It then generates an internal ``thought'' to reason about the state and produce another ``action'' in the form of a tool invocation. 
This observation-thought-action cycle repeats, allowing the model to iteratively complete the design.
The process terminates when the model stops invoking a tool invocation, signifying that it views the task as complete.

\subsection{Model Configuration and Hyperparameters}
We document the core hyperparameters for the models used in our benchmark. All models use a fixed temperature of $0.0$ to ensure deterministic behavior, unless otherwise noted. Token and turn limits vary depending on the task type and agent architecture.

\begin{table}[h]
\centering
\caption{Hyperparameters by Variant for Replication Task}
\label{tab:replication-hparams}
\begin{adjustbox}{max width=\columnwidth}
\small
\begin{tabular}{llcc}
\toprule
\textbf{Architecture} & \textbf{Model} & \textbf{Max Tokens} & \textbf{Max Turns} \\
\midrule
\multirow[c]{5}{*}{\parbox{1.7cm}{\centering Tool\\(Agent)}}
  & GPT-4o & 2048 & 50 \\
  & GPT-4.1 & 2048 & 50 \\
  & Claude-3.5-Sonnet & 2048 & 50 \\
  & Gemini-2.5-Flash & 2048 & 50 \\
  & Gemini-2.5-Pro & 2048 & 50 \\
\midrule
\multirow{5}{*}{\shortstack{Tool\\(Single-Turn)}} 
  & GPT-4o & 2048 & 1 \\
  & GPT-4.1 & 2048 & 1 \\
  & Claude-3.5-Sonnet & 2048 & 1 \\
  & Gemini-2.5-Flash & 2048 & 1 \\
  & Gemini-2.5-Pro & 2048 & 1 \\
\midrule
\multirow{5}{*}{\shortstack{Code\\(Single-Turn)}}
  & GPT-4o & 4096 & 1 \\
  & GPT-4.1 & 32768 & 1 \\
  & Claude-3.5-Sonnet & 8192 & 1 \\
  & Gemini-2.5-Flash & 65536 & 1 \\
  & Gemini-2.5-Pro & 65536 & 1 \\
\bottomrule
\end{tabular}
\end{adjustbox}
\end{table}

\begin{table}[h]
\centering
\caption{Hyperparameters by Variant for Modification Task}
\label{tab:modification-hparams}
\begin{adjustbox}{max width=\columnwidth}
\small
\begin{tabular}{llcc}
\toprule
\textbf{Architecture} & \textbf{Model} & \textbf{Max Tokens} & \textbf{Max Turns} \\
\midrule
\multirow[c]{5}{*}{\parbox{1.7cm}{\centering Tool\\(Agent)}}
  & GPT-4o & 2048 & 50 \\
  & GPT-4.1 & 2048 & 50 \\
  & Claude-3.5-Sonnet & 2048 & 50 \\
  & Gemini-2.5-Flash & 2048 & 50 \\
  & Gemini-2.5-Pro & 2048 & 50 \\
\bottomrule
\end{tabular}
\end{adjustbox}
\end{table}

\begin{table}[H]
\centering
\caption{Model Version and Access Method for Tool Invocation}
\label{tab:model-access}
\begin{adjustbox}{max width=\columnwidth}
\small
\begin{tabular}{lll}
\toprule
\textbf{Model} & \textbf{Provider} & \textbf{Model Version / API ID} \\
\midrule
GPT-4o & OpenAI API & \texttt{gpt-4o-2024-08-06} \\
GPT-4.1 & OpenAI API & \texttt{gpt-4.1-2025-04-14} \\
Claude 3.5 Sonnet & Amazon Bedrock & 
\makecell[l]{\texttt{anthropic.claude-3-5}\\\texttt{-sonnet-20241022-v2:0}} \\
Gemini 2.5 Flash & Google AI Studio & \texttt{gemini-2.5-flash} \\
Gemini 2.5 Pro & Google AI Studio & \texttt{gemini-2.5-pro} \\
\bottomrule
\end{tabular}
\end{adjustbox}
\end{table}

For collecting a tool invocation, we utilized the official parameter that invokes the tool.
For the GPT model family, we referred to function calling in the Response API.
For the Claude model, we referred to the tool use in the  Message API.
For the Gemini model family, we referred to the function calling in the Gemini API.
Note that for the Gemini models, we had the ``thinking'' on following the base setting, which sets the mode as ``dynamic thinking.''

\clearpage

\clearpage

\onecolumn
\begin{small}
\setlength{\tabcolsep}{4pt}
\renewcommand{\arraystretch}{0.92}
\begin{table*}[htbp]
  \centering
  \caption{List of 50 Tools Included in the \benchname{}}
  \scriptsize
  \begin{tabular}{@{}p{5.0cm}p{11.3cm}@{}}
    \toprule
    \textbf{Category \& Tool} & \textbf{Description}\\
    \midrule

    \multicolumn{2}{@{}l}{\textbf{Connection}}\\
    \addlinespace
    \quad \texttt{get\_channels}             & Return a communication channel array currently open for environment communication.\\
    \quad \texttt{select\_channel}           & Select a communication channel.\\
    \quad \texttt{check\_connection\_status} & Check the environment to verify whether the communication is live.\\
    \midrule

    \multicolumn{2}{@{}l}{\textbf{Creation}}\\
    \addlinespace
    \quad \texttt{create\_rectangle}  & Create a rectangle with width, height, fill \& stroke along with other attributes.\\
    \quad \texttt{create\_frame}      & Create an auto-layout frame with padding, spacing, resize rules, and style props.\\
    \quad \texttt{create\_text}       & Create a text node with content string, font family, size, alignment, and color token.\\
    \quad \texttt{create\_graphic}    & Create vector nodes based on the SVG markup code text.\\
    \quad \texttt{create\_ellipse}    & Create an ellipse with configurable radii, fill, stroke, and corner smoothing.\\
    \quad \texttt{create\_polygon}    & Create an \(N\)-sided polygon with adjustable corner radius, fill/stroke, and rotation.\\
    \quad \texttt{create\_star}       & Create a star with custom points and inner-radius.\\
    \quad \texttt{create\_line}       & Create a straight line between two absolute points with dash pattern and end-caps props.\\
    \quad \texttt{create\_mask}       & Create a mask node using group of nodes to create a clipping group.\\
    \midrule

    \multicolumn{2}{@{}l}{\textbf{Inspection}}\\
    \addlinespace
    \quad \texttt{get\_page\_info}            & Return list of top-level nodes on the page with names, IDs, types, and positions.\\
    \quad \texttt{get\_selection\_info}       & Return full property sets (geometry, styles) for nodes in the active selection.\\
    \quad \texttt{get\_node\_info}            & Return metadata, bounding boxes, style refs, and hierarchy path given the node IDs.\\
    \quad \texttt{get\_node\_info\_by\_types} & Traverse a parent and return children information that match the specified type.\\
    \quad \texttt{get\_result\_image}         & Rasterise the visible design to PNG/JPEG and return an encoded image blob.\\
    \quad \texttt{get\_page\_structure}       & Return a full node tree of the page, including order and absolute transforms.\\
    \quad \texttt{export\_json}               & Return the entire page as a structured JSON format.\\
    \quad \texttt{import\_json}               & Reconstruct a design on the page from a serialized JSON structure string.\\
    \midrule

    \multicolumn{2}{@{}l}{\textbf{Layout}}\\
    \addlinespace
    \quad \texttt{set\_padding}        & Set per-side padding inside an auto-layout frame for internal spacing control.\\
    \quad \texttt{set\_axis\_align}    & Set child alignment on primary/cross axes.\\
    \quad \texttt{set\_layout\_sizing} & Set width/height mode of the layout.\\
    \quad \texttt{set\_item\_spacing}  & Set uniform gap distance between auto-layout children.\\
    \quad \texttt{set\_layout\_mode}   & Set layout direction and optional wrapping.\\
    \midrule

    \multicolumn{2}{@{}l}{\textbf{Operation}}\\
    \addlinespace
    \quad \texttt{move\_node}      & Translate a node to new \(x,y\) coordinates and optionally re-parent it.\\
    \quad \texttt{clone\_node}     & Duplicate a node, retaining constraints and styles in the copy.\\
    \quad \texttt{resize\_node}    & Set width/height while preserving position or constraints.\\
    \quad \texttt{delete\_node}    & Remove one or more nodes from the canvas and history stack.\\
    \quad \texttt{group\_nodes}    & Group selected nodes into a group layer.\\
    \quad \texttt{ungroup\_nodes}  & Dissolve a group, restoring its children to the parent container.\\
    \quad \texttt{rename\_node}    & Change the name of a node for better organization.\\
    \quad \texttt{rotate\_node}    & Apply rotation in degrees based on the node centre.\\
    \quad \texttt{boolean\_nodes}  & Perform a boolean operation to create a composite node.\\
    \quad \texttt{reorder\_node}   & Move node forward/backward in sibling z-order.\\
    \midrule

    \multicolumn{2}{@{}l}{\textbf{Style}}\\
    \addlinespace
    \quad \texttt{set\_fill\_color}    & Set a solid color fill.\\
    \quad \texttt{set\_corner\_radius} & Set uniform or per-corner radii to round a node’s corners.\\
    \quad \texttt{get\_styles}         & Return arrays of all shared text, color, and effect styles in the file.\\
    \quad \texttt{set\_opacity}        & Set overall opacity (0–100\%) of the component.\\
    \quad \texttt{set\_stroke}         & Set stroke color, weight, alignment, and dash pattern.\\
    \quad \texttt{set\_fill\_gradient} & Set linear, radial, angular, or diamond gradient with multi-stop control.\\
    \quad \texttt{set\_drop\_shadow}   & Add a drop shadow specifying color, blur radius, offset, and spread.\\
    \quad \texttt{set\_inner\_shadow}  & Add an inner shadow confined within node bounds.\\
    \quad \texttt{copy\_style}         & Copy all visual properties from a source node to target nodes.\\
    \quad \texttt{set\_blend\_mode}    & Set blend mode for compositing with layers beneath.\\
    \midrule

    \multicolumn{2}{@{}l}{\textbf{Text}}\\
    \addlinespace
    \quad \texttt{set\_text\_content}    & Replace the text string of one or many text nodes.\\
    \quad \texttt{get\_text\_node\_info} & Return typography \& content data for each text node in scope.\\
    \quad \texttt{set\_text\_properties} & Set font size, line height, letter spacing, alignment, and paragraphs.\\
    \quad \texttt{set\_text\_decoration} & Set underline, strikethrough, superscript, or case transformations.\\
    \quad \texttt{set\_text\_font}       & Set font family and style weight; auto-loads missing fonts if present.\\
    \bottomrule
  \end{tabular}
  \label{tab:tool-list}
\end{table*}
\end{small}
\twocolumn

\subsection{Figma Environment and Tool List}

All tools available to the VLM are implemented using the official Figma Plugin API. 
When a tool is invoked, it executes a command on the Figma canvas and returns a result in a descriptive text and structured JSON format. 
This result reports the outcome of the operation, such as its success or failure, or provides any data queried from the environment.

The toolset is organized into seven comprehensive categories that cover the primary design operations available in Figma: \texttt{connection}, \texttt{creation}, \texttt{inspection}, \texttt{layout}, \texttt{operation}, \texttt{style}, and \texttt{text}.

\begin{table}[h]
\small                 
\caption{High-level Tool Categories}
\centering
\begin{tabularx}{\columnwidth}{lX}
\toprule
\textbf{Category} & \textbf{Purpose} \\
\midrule
\texttt{Connection} & Establishes and manages the connection to the Figma environment. \\[2pt]

\texttt{Creation}   & Creates new design elements, such as rectangles, frames, and other objects. \\[2pt]

\texttt{Inspection} & Queries the current state of the design canvas, including the properties of existing elements. \\[2pt]

\texttt{Layout}     & Adjusts the alignment, distribution, and auto-layout properties of target components. \\[2pt]

\texttt{Operation}  & Manipulates the hierarchy, position, and grouping of design elements. \\[2pt]

\texttt{Style}      & Controls the visual appearance and aesthetic properties of components, including their fills, strokes, and effects. \\[2pt]

\texttt{Text}       & Edits textual content and modifies text-specific styling properties like font, size, and weight. \\
\bottomrule
\end{tabularx}
\end{table}

To increase the reliability of our benchmark, we intentionally excluded features designated as ``beta'' in the Figma Plugin API documentation, such as advanced texture and blur effects.
Table~\ref{tab:tool-list} shows the full list of the tools included in the \benchname{}.

\subsubsection{Prompt Design for the Replication and Modification Task}

We employ a modular design for our task instruction prompts to ensure coherent context across all tasks. 
This approach involves two common instruction blocks, \texttt{Figma Tool Basics} and \texttt{Agency Principle}, which are added to each task-specific prompt as a variable.

The \texttt{Figma Tool Basics} block, shown in Figure~\ref{fig:figma-tool-basics-prompt}, outlines the VLM with basic knowledge about the Figma environment and general principles for structuring a design hierarchy. 
Based on findings from our early experiments, we identified that incorrect handling of layout and text elements was the most frequent source of task failure. 
Therefore, this block includes explicit guidelines on these topics to mitigate common errors and prevent significant performance degradation.

\begin{figure}[h]
  \centering
  \caption{\textbf{Figma Tool Basics Prompt:} A prompt for guiding models on the use of the tools for producing a design in the Figma environment.}
  \begin{lstlisting}[
      basicstyle=\ttfamily\scriptsize,
      numberstyle=\tiny,
      frame=single,
      numbers=left,
      breakindent=0pt
    ]
1. Figma Tool Basics
- In Figma tool calls, each design element appears as a node representing either a container (frame/component) or a leaf (shape/text).
- Nodes provide uniform structural data exposing their unique properties.
- Coordinates are global: all nodes sit relative to the canvas origin (0, 0) at the top-left.

2. Node Hierarchy
- All nodes live in one rooted tree mirroring the layer list.
- Parent-child links create the hierarchy, and a node's index in its parent sets both z-order and sidebar order.
- When child nodes (leaf) outgrow their parent nodes (container), they will be clipped.

3. Container Layout
- With auto layout applied to the frame, Figma automatically manages direction, gap, padding, and resizing of the container and the children.
- So, manual layout property changes must account for these automatic adjustments of size and position.
- Enable auto layout only when confident, as it can cause unexpected shifts.

4. Text Mechanics
- Text nodes expose font family, style, size, and other typography traits independent of layout.
- Resizing the text node doesn't scale the text; excess text simply overflows.
- Adequately set the text node size and alignment to avoid overflow.
  \end{lstlisting}
  \label{fig:figma-tool-basics-prompt}
\end{figure}

The \texttt{Agency Principle} instruction block, shown in Figure~\ref{fig:agency-principle-prompt}, defines the operational protocol for the VLM to function as an autonomous agent. 
It instructs the model to complete the UI design task through an iterative process of reasoning and tool use. 
A primary goal of this prompt is to prevent the model from prematurely terminating the task, thereby encouraging persistent, multi-step problem-solving.

\begin{figure}[h]
  \centering
  \caption{\textbf{Agency Principle Prompt:} A prompt for guiding models on the behavior as an agent when interacting with the environment.}
  \begin{lstlisting}[
      basicstyle=\ttfamily\scriptsize,
      numberstyle=\tiny,
      frame=single,
      numbers=left,
      breakindent=0pt
    ]
1. Persistence  
Keep iterating until the instruction is fully met and confirmed. Do not end the turn early.
2. Tool use  
Interact with the canvas via the provided Figma-control tools.
3. Keen Examination
Carefully examine the instructions and image (if provided) and follow them accordingly.
  \end{lstlisting}
  \label{fig:agency-principle-prompt}
\end{figure}

The instruction prompt for the Replication Task, shown in Figure~\ref{fig:replication-task-prompt}, directs the model to reproduce a target design from a ground-truth image. 
The prompt incorporates our standard \texttt{Agency Principle} and \texttt{Figma Basics} blocks to provide the necessary operational and environmental context. 
To further aid the models in spatial operations, the prompt also specifies the target frame dimensions (width and height), which define the coordinate system for the model.

For the reader's clarity when reviewing the figure, we note a minor terminological inconsistency in the prompt text: the \texttt{Figma Basics} block is referred to as \texttt{UI Design Principles}.

\begin{figure}[h]
  \centering
  \caption{\textbf{Design Replication Prompt:} A prompt for replicating a UI design. The ``\$\{\}'' indicates a variable.}
  \begin{lstlisting}[
      basicstyle=\ttfamily\scriptsize,
      numberstyle=\tiny,
      frame=single,
      numbers=left,
      breakindent=0pt
    ]
**Context**
You are a UI-design agent with access to Figma via tool calls.
Follow the **Instruction** to generate a UI design.
Refer to the **Agency Principles** and **UI Design Principles** for guidance.

**Agency Principles**
${agencyPrinciples}

**Figma Basics**
${figmaInstruction}

**Instruction**
Please analyze the following image and reproduce the UI design inside the existing "Main Screen" frame in the Figma, exactly.
The frame size is ${width}x${height} pixels.
  \end{lstlisting}
  \label{fig:replication-task-prompt}
\end{figure}

The instruction prompt for the Modification Task, shown in Figure~\ref{fig:modification-task-prompt}, is largely identical to that of the Replication Task. 
The key distinction is the inclusion of an additional \texttt{Text} field, which contains the specific natural language instruction detailing the required design modification.

\begin{figure}[h]
  \centering
  \caption{\textbf{Design Modification Prompt:} A prompt for modfiying a UI design. The ``\$\{\}'' indicates a variable.}
  \begin{lstlisting}[
      basicstyle=\ttfamily\scriptsize,
      numberstyle=\tiny,
      frame=single,
      numbers=left,
      breakindent=0pt
    ]
**Context**
You are a UI-design agent with access to Figma via tool calls.
Follow the **Instruction** to modify a UI design.
Refer to the **Agency Principles** and **Figma Basics** for guidance.

**Agency Principles**
${agencyPrinciples}

**Figma Basics**
${figmaInstruction}

**Instruction**
Please analyze the provided screen image and text instruction, then update the UI design within the existing "Main Screen" frame in Figma to precisely match the image and the instruction.
The frame size is ${width}x${height} pixels.
Text: ${instruction}
  \end{lstlisting}
  \label{fig:modification-task-prompt}
\end{figure}

\subsubsection{Prompt Design for the Single-Turn Tool-based Generation and Code-based Generation Task}

For our ablation study, we designed specialized prompts for two additional conditions: single-turn tool-based generation and code-based generation. 
Both of these conditions were evaluated exclusively on the Replication Task.

For the single-turn tool-based generation condition, we modified the standard prompt by replacing the \texttt{Agency Principle} block with a new \texttt{Tool Use Principle} block, shown in Table~\ref{fig:single-turn-generation-prompt}. 
The new prompt instructs the model to forgo the iterative agentic loop. 
Instead, its objective is to generate a single, exhaustive tool invocation capable of constructing the entire target design in one turn.

\begin{figure}[h]
  \centering
  \caption{\textbf{Tool-Use Principle Prompt:} A prompt for outlining the guidance for the single-turn tool-based generation.}
  \begin{lstlisting}[
      basicstyle=\ttfamily\scriptsize,
      numberstyle=\tiny,
      frame=single,
      numbers=left,
      breakindent=0pt
    ]
Your task is to produce an array of tool (function) calls necessary to recreate the design in one turn.
Include every necessary tool (function) calls and do not output any text other than the function calls themselves.

1. Plan first
Briefly outline the key steps you will take.
2. Be exhaustive
Consider all parameters, options, ordering, and dependencies necessary to recreate the design in one turn.
3. Deliver
Based on the plan, respond with the exact sequence of tool(function) calls it should run.
  \end{lstlisting}
  \label{fig:single-turn-generation-prompt}
\end{figure}

For the code-based generation condition, our prompt is based on the methodology from Design2Code (Si et al., 2024) to closely approximate their experimental setup. The full prompt is shown in Table~\ref{fig:code-based-generation-prompt}.
While we reproduced the core structure of the original prompt, we introduced two modifications to integrate it into our framework:
We instructed the model to wrap the generated HTML/CSS code in triple backticks to ensure reliable parsing of the output.
We explicitly predefined the output canvas dimensions (width and height in pixels). 
This modification ensures a fair comparison with our tool-based methods, where screen dimension information is also provided to assist with spatial operations.

\begin{figure}[h]
  \centering
  \caption{\textbf{Code-Based Generation Prompt:} A prompt for replicating a UI design using code. The ``\$\{\}'' indicates a variable.}
  \begin{lstlisting}[
      basicstyle=\ttfamily\scriptsize,
      numberstyle=\tiny,
      frame=single,
      numbers=left,
      breakindent=0pt
    ]
You are an expert web developer who specializes in HTML and CSS. A user will provide you with a screenshot of a mobile app.
You need to return a single html file that uses HTML and CSS to reproduce the given mobile app.
Include all CSS code in the HTML file itself. If it involves any images, use "rick.jpg" as the placeholder.
Some images on the webpage are replaced with a blue rectangle as the placeholder, use "rick.jpg" for those as well.
Do not hallucinate any dependencies to external files.
You do not need to include JavaScript scripts for dynamic interactions. Pay attention to things like size, text, position, and color of all the elements, as well as the overall layout.
Respond with the content of the HTML+CSS file. Wrap the code in backticks.
The page must be designed to match ${width}-pixel width and ${height}-pixel height.
  \end{lstlisting}
  \label{fig:code-based-generation-prompt}
\end{figure}

\subsubsection{Modification Task Prompt Examples}

The modification task consists of three sub-tasks: \texttt{attribute update}, \texttt{component insertion}, and \texttt{mode change}.
For \texttt{attribute update} and \texttt{component insertion}, each task instance is paired with a unique, case-specific instruction prompt. 
To illustrate these instructions, we include representative examples from the dataset below.
Each example presents two images depicting the initial (base) state and the final (target) state, along with the corresponding text instruction.
For every subtask, the target image is provided to the model during inference to ensure the model approximates the ground truth result.

\begin{figure}[h]
  \centering
  \includegraphics[width=\columnwidth]{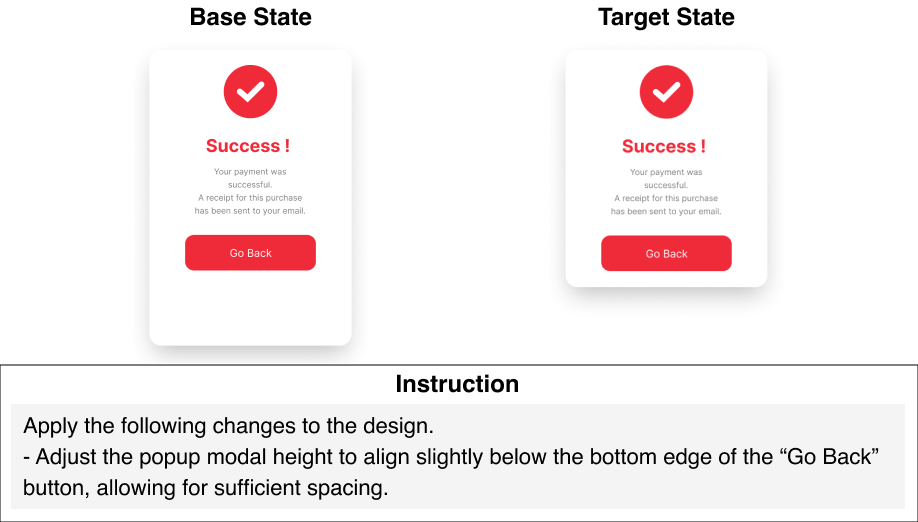}
  \caption{\textbf{Modification Task-1 Instruction Example} An example task case from the Modification Task-1, \emph{attribute update}, with the instruction.}
  \label{fig:modification-task-one-example}
\end{figure}

\begin{figure}[H]
  \centering
  \includegraphics[width=\columnwidth]{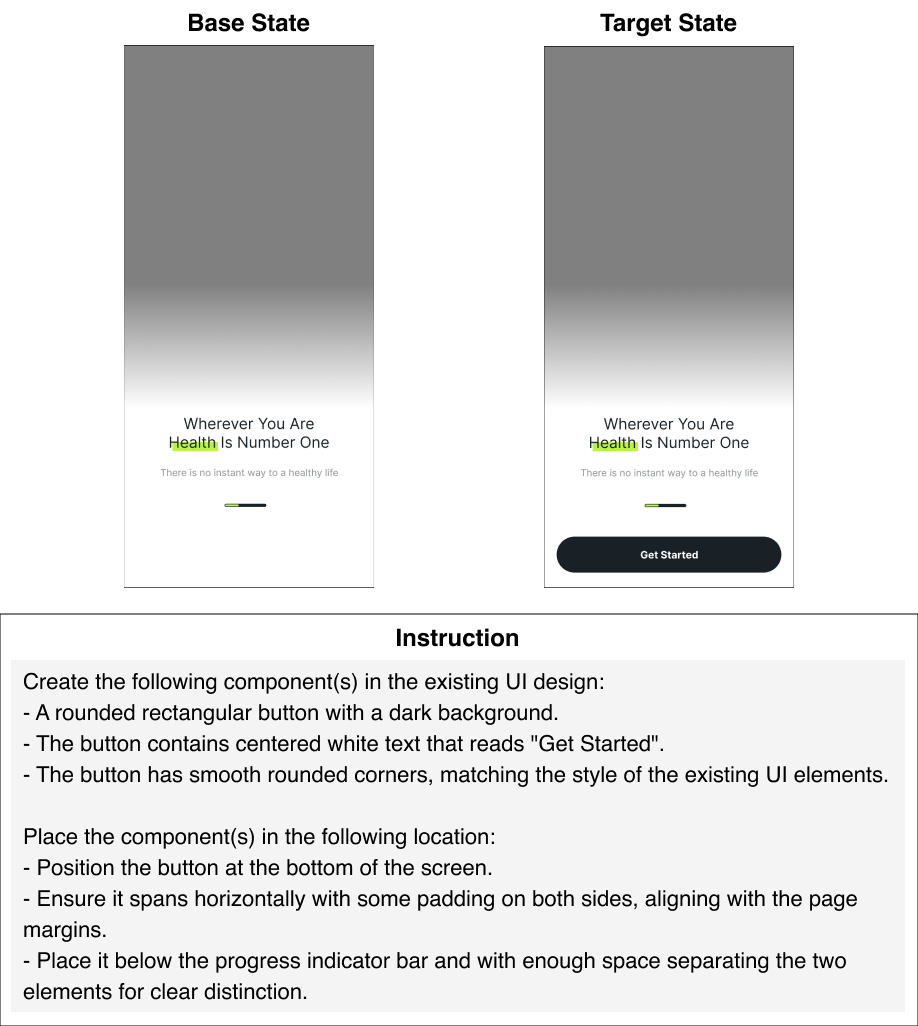}
  \caption{\textbf{Modification Task-2 Instruction Example} An example task case from the Modification Task-2, \emph{component insertion}, with the instruction.} 
  \label{fig:modification-task-two-example}
\end{figure}

\begin{figure}[H]
  \centering
  \includegraphics[width=\columnwidth]{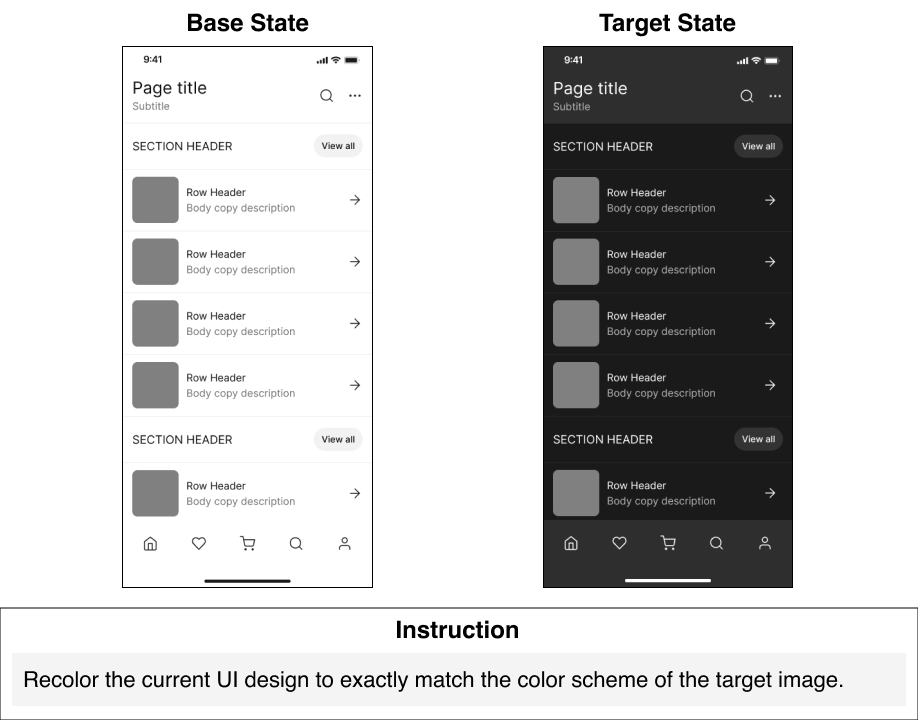}
  \caption{\textbf{Modification Task-3 Instruction Example} An example task case from the Modification Task-3, \emph{mode change}, with the instruction.}
  \label{fig:modification-task-three-example}
\end{figure}

\begin{itemize}
    \item \textbf{Attribute Change:} This sub-task requires the model to modify a property of an existing element. As illustrated in Figure~\ref{fig:modification-task-one-example}, the model is provided with the base image and a case-specific text instruction. 
    These instructions intentionally use qualitative, relational descriptions (e.g., resizing a ``popup modal'' to be ``slightly below the bottom edge of the `Go Back' button'') rather than precise numerical values. 
    This design choice aims to reflect real-world scenarios where designers communicate changes colloquially. 
    \item \textbf{Component Insertion:} For this sub-task, the model must add a new element to the design. 
    The instruction, also unique to each case (Fig.~\ref{fig:modification-task-two-example}), specifies the required visual properties, internal structure, and location of the new component relative to the existing design components.
    \item \textbf{Mode Change:} Unlike the other two sub-tasks, \texttt{mode change} utilizes a single, generic instruction for all instances in its set (Fig.~\ref{fig:modification-task-three-example}). 
    This approach is used because the required transformation is uniform across all cases, which involves applying a color palette shift to toggle between light and dark modes.
\end{itemize}

\clearpage

\section{Metric Implementation Details}

We evaluate model performance by adapting established metrics from code-based UI generation to tool-based generation, as both produce outputs with associated image and tree-based representations (HTML or JSON). 
However, as discussed in the \textit{Related Work} section, structural mismatches between code-based outputs and tool-generated designs make direct reuse of existing metrics infeasible. 
To address this, we adapt the evaluation protocol to the newly suggested design environment, aligning the structured outputs (exported as JSON) from model generations with those of the ground-truth designs for comparison.

Drawing inspiration from the theory of human visual processing, we propose a hierarchical metric framework designed for human-centric interpretation of the evaluation results.
The framework introduces two metric categories: (i) \textit{perceptual similarity}, which measures similarity in the context of human visual perception; and (ii) \textit{component-wise similarity}, which evaluates similarity at an individual component level.

\subsection{Perceptual Similarity Metrics}

\benchname{} decomposes the visual similarity in design into three hierarchical levels based on the theory of the human bottom-up visual processing: features, patterns, and objects~\cite{ware2010visual}.
The hierarchy explains how a human sequentially extracts a meaning from an image, assembling visual features (e.g., edges) into patterns (e.g., a rectangular shape), and inferring a semantic object from it (e.g., a button).
By mapping the metrics to three levels, the benchmark adds interpretability to contrasting trends between metrics within the same design.
Precisely, we adopt algorithmic techniques designed to approximate visual characteristics that closely correspond to each stage of processing.

\begin{itemize}
\item \emph{\textbf{Structural Similarity Index (SSIM)}}:
The SSIM captures mid-level structural patterns by comparing local means $\mu$, variances $\sigma^{2}$, and cross–covariance $\sigma_{xy}$ between two image patches. 
Hence, it exposes the low-level composition of pixels that build the structure of a shape (e.g., size and orientation) in design, corresponding to the feature.
We calculate $\mathrm{SSIM}(\cdot)$ using the \texttt{scikit-image} library with a 7×7 Gaussian window (default setting) and channel-wise averaging (\texttt{multichannel=True}).
$$\mathrm{SIM}_{\text{feat}}(s_t,s_{GT})=\mathrm{SSIM}(I_t,I_{GT})$$
where $I_t$ and $I_{GT}$ denote the rasterized images of $s_t$ and $s_{GT}$, respectively.

\item \emph{\textbf{Saliency Map Similarity}}: The saliency map estimates human attention based on the composition of low-level visual features, highlighting regions that may guide focus within a UI design.
We generate saliency maps $\Phi(\cdot)$ using the pretrained model named \texttt{UMSI++}~\cite{jiang2023ueyes}, resize both ground-truth and generated images to 320×240, and compute pattern-level similarity via the Pearson Correlation Coefficient (CC($\cdot$)) between the two maps.
Each saliency map is first min-max scaled to $[0,1]$ and then standardized (zero mean, unit variance) before comparison. Inference is performed in \texttt{eval()} mode with all stochastic layers disabled and the random seed fixed to 42 for reproducibility. 
$$
\mathrm{SIM}_{\text{pat}}(s_t, s_{GT}) = \mathrm{CC}(\Phi(I_t), \Phi(I_{GT}))
$$

\begin{figure}[h]
  \centering
  \includegraphics[width=\columnwidth]{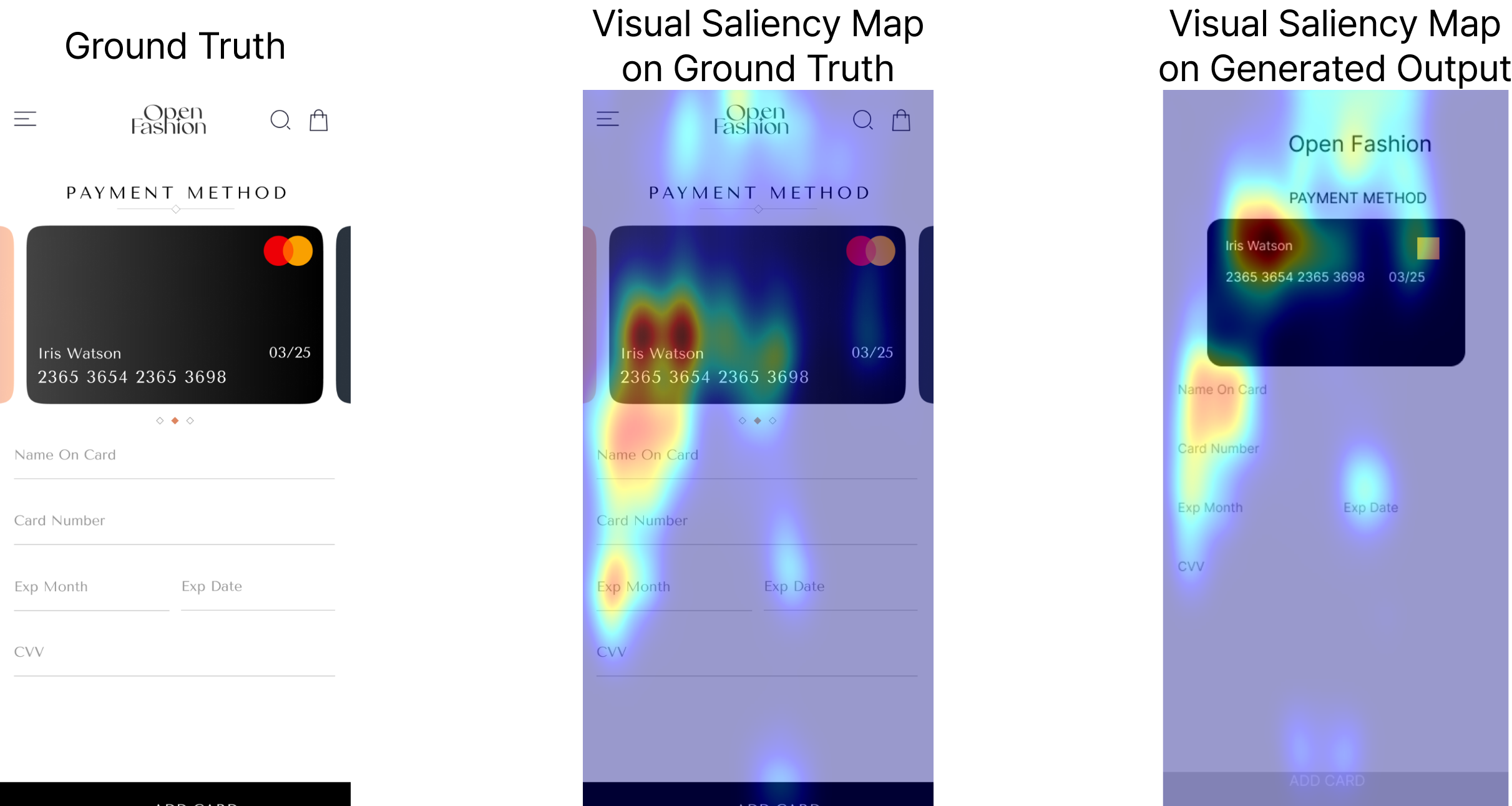}
  \caption{\textbf{Visual saliency comparison} The middle and right panels show inferred saliency overlays on the ground truth and generated UIs, respectively. Similarity scores (CC = 0.5419) quantify the alignment between human and model attention.}

  \label{fig:visual-saliency-example}
\end{figure}

\item \emph{\textbf{BLIPScore}}: The BLIP caption approximates object-level semantics by generating textual descriptions of a design image, which reflect how humans may semantically interpret its content.
We generate captions using the \texttt{Salesforce/blip2-opt-2.7b} model with greedy decoding (\texttt{num\_beams=1}, \texttt{do\_sample=False}).
\texttt{sentence-transformers/all-MiniLM-L6-v2} is used to embed the resulting texts.
All inference steps are conducted in \texttt{eval()} mode with the random seed fixed to 42 to ensure deterministic outputs.
Semantic similarity is then computed as the cosine similarity between the caption embeddings:
$$
\mathrm{SIM}_{\text{obj}}(s_t, s_{GT}) = 
\frac{\langle\,\psi(I_t),\,\psi(I_{GT})\,\rangle}
     {\left\lVert \psi(I_t) \right\rVert \cdot \left\lVert \psi(I_{GT}) \right\rVert}
$$
where $\psi(\cdot)$ denotes the embedding of the generated caption for the given image.
\end{itemize}

\subsection{Component-wise Similarity Metrics}
We adapt the component-level matching principles from prior works ~\cite{si2024design2code, li2024sketch2code}, which evaluate how well individual UI elements (e.g., text, buttons, containers) are reproduced in the generated output. 
Unlike static input settings (e.g., sketches or images), our task involves tool-based, multi-turn generation within a design environment, where outputs are structured and semantically labeled. 
Accordingly, we reinterpret component-wise similarity in the context of tool-generated Figma designs by leveraging structured metadata (e.g., bounding boxes, fill styles, characters) for fine-grained element matching. 
This preserves the per-element evaluation spirit of Design2Code, while aligning with our scenario where models construct editable, design-native artifacts.

The evaluation comprises two stages: (i) Component Matching via bounding alignment and and (ii) Attribute Similarity calculation across matched pairs. 
The resulting similarity score reflects structural and semantic consistency between generated and ground-truth states.

\subsubsection{Component Matching}
Our matching strategy treats text and non-text components separately due to their distinct properties. Let $\mathcal{C}_{\text{GT}}$ and $\mathcal{C}_{\text{t}}$ denote the sets of ground-truth and generated components, respectively.

\paragraph{Non-text components}
We compute the Intersection over Union (IoU) between the ground-truth and generated boxes $B$. The cost $C$ is defined as:

$$
C_{ij} = 1 - \text{IoU}(B_{t}, B_{GT})
$$ Only matches above an IoU threshold (e.g., 0.5) are retained.

\paragraph{Text components}
Text boxes are matched using a composite similarity score combining:
\begin{itemize}
    \item \textbf{Text similarity}($S_{\text{text}}$): computed using a mix of length ratio and character-level Jaccard similarity.
    \item \textbf{Position similarity}($S_{\text{pos}}$): defined by the L-infinity distance between box centers.
\end{itemize}

\noindent The overall cost is computed as a weighted combination:

$$
C_{ij} = \alpha \cdot (1 - S_{\text{text}}) + \beta \cdot (1 - S_{\text{pos}})
$$ with equal weights \(\alpha = \beta = 0.5\) by default.

\paragraph{Block Match}
The cost matrices $C_{ij}$ for text and non-text components are used as input to the Hungarian algorithm, which computes one-to-one assignments between $\mathcal{C}_{\text{GT}}$ and $\mathcal{C}_{\text{t}}$ for each group. The resulting matched pairs are combined into a unified set $\mathcal{M}^\star$.
The Block Match Score is then defined as:
\begin{equation}
\mathrm{S}_{\text{match}} = \frac{|\mathcal{M}^\star|}{|\mathcal{C}_{\text{GT}}|}.
\tag{1}
\end{equation}

\subsubsection{Similarity Metrics Calculation}

We compute three attribute-level similarity metrics based on the matched component pairs $\mathcal{M}^\star$ between $\mathcal{C}_{\text{GT}}$ and $\mathcal{C}_{\text{t}}$. All metrics are averaged over the matched pairs and capture different aspects of visual similarity.

\paragraph{Position Similarity}
For each $(i,j) \in \mathcal{M}^\star$, we compute the Euclidean distance between the centers of the bounding boxes and normalize it by the larger diagonal of the two boxes:
\begin{equation}
\mathrm{SIM}_{\text{pos}} = \frac{1}{|\mathcal{M}^\star|} \sum_{(i,j) \in \mathcal{M}^\star} \left( 1 - \frac{\| c_i - c_j \|_2}{\max(\text{diag}_i, \text{diag}_j)} \right)
\tag{2}
\end{equation}

\paragraph{Color Similarity}
For matched pairs that both have solid fill colors, we compute the normalized Euclidean distance between their RGB values:
\begin{equation}
\mathrm{SIM}_{\text{col}} = \frac{1}{|\mathcal{M}_{\text{color}}^\star|} \sum_{(i,j) \in \mathcal{M}_{\text{color}}^\star} \left( 1 - \frac{\| \text{rgb}_i - \text{rgb}_j \|_2}{\sqrt{3} \cdot 255} \right)
\tag{3}
\end{equation}
where $\mathcal{M}_{\text{color}}^\star \subseteq \mathcal{M}^\star$ includes only pairs where both components define a solid color.

\paragraph{Text F1 Score}
Unlike the above metrics, text similarity is computed independently of component matching. We extract all text strings from $s_{\text{GT}}$ and $s_{\text{new}}$ into multisets $\mathcal{T}_{\text{GT}}$ and $\mathcal{T}_{\text{t}}$, respectively, and compute a global F1 score over string occurrences:
\begin{equation}
\mathrm{SIM}_{\text{text}} = \mathrm{F1}(\mathcal{T}_{\text{t}}, \mathcal{T}_{\text{GT}})
\tag{4}
\end{equation}
This metric evaluates whether the correct text contents are present in the generated design, regardless of their positions or bounding boxes.

\paragraph{Interpretation}
Position and color similarity are computed only over successfully matched components in $\mathcal{M}^\star$, and thus reflect how well aligned or visually faithful each matched pair is. In contrast, text similarity assesses global content reproduction by comparing the sets of texts used, irrespective of spatial location or matching

\subsubsection{Component-wise Similarity} 
We compute the overall component-wise similarity by averaging the four component-level scores introduced above: Block Match Score, Position Similarity, Color Similarity, and Text F1 Score. Formally, the final component-wise score is calculated as:
\begin{align*}
\mathrm{SIM}_{\text{comp}}(s_t,s_{GT}) 
&= \frac{1}{4} \big( 
     \mathrm{S}_{\text{match}} + \mathrm{SIM}_{\text{pos}} \\
&\quad \quad \ + \mathrm{SIM}_{\text{col}} + \mathrm{SIM}_{\text{text}} \big)
\end{align*}
where all component-level metrics are computed between the generated design $s_t$ and the ground-truth design $s_{\text{GT}}$. This unified score reflects both structural and content-level consistency between the generated and ground-truth UI designs.

\subsection{Metric Aggregation Protocol}

For each benchmark case, we compute similarity scores between the generated design $s_t$ and the ground-truth design $s_{GT}$ using a family of metric functions $\mathrm{SIM}_m(s_t, s_{GT})$, where $m \in \{\text{SSIM}, \text{Saliency}, \text{BLIP}, \text{Component-wise}\}$. Then, for each metric $m$, we report the model-level score as the average over all test cases:
$$
\mathrm{Score}_m = \frac{1}{N} \sum_{i=1}^N \mathrm{SIM}_m\left(s_t^{(i)}, s_{GT}^{(i)}\right)
$$

\subsubsection{Replication} The model generates a design $s_t$ from scratch using only the reference image $s_{GT}$ in the replication task. We report $\mathrm{Score}_m$ along with its standard deviation across all examples, denoted by $\pm$.

\subsubsection{Modification} The model transforms an initial state $s_{old}$ into an updated design $s_{new}$, targeting the desired state $s_{GT}$ in the modification task. We evaluate each metric by comparing $s_{new}$ to $s_{GT}$, and further compute the score improvement relative to the base design $s_{old}$:
$$
\Delta_m = \mathrm{SIM}_m(s_{new}, s_{GT}) - \mathrm{SIM}_m(s_{old}, s_{GT})
$$
where the reported $\mathrm{Score}_m$ reflects the similarity between $s_{new}$ and $s_{GT}$. $\Delta$ denotes the improvement from the base design.

\begin{figure*}[!t]
  \centering
  \includegraphics[width=\textwidth]{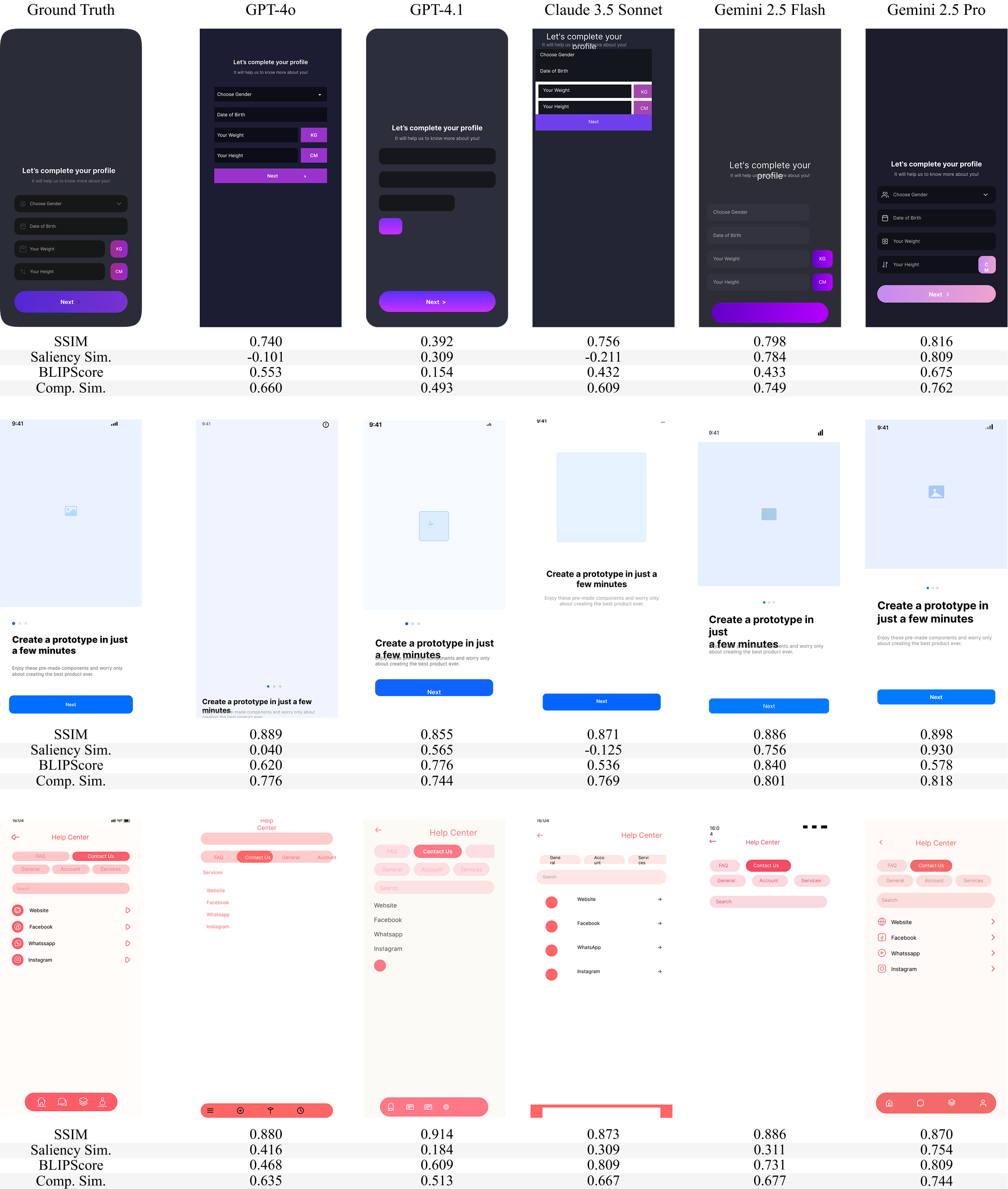} 
\caption{\textbf{Qualitative results with evaluation scores for the replication task.} Each row shows one example (\texttt{gid-51-43}, \texttt{gid-1-11}, \texttt{gid-78-11}), comparing model outputs to the ground truth. Reported metrics include SSIM, saliency map similarity (Saliency Sim.), BLIPScore, and component-wise similarity (Comp. Sim.). Higher values indicate greater similarity to the ground truth.}
  \label{fig:qual_replication}
\end{figure*}

\begin{figure*}[t]
  \centering
  \includegraphics[width=\textwidth]{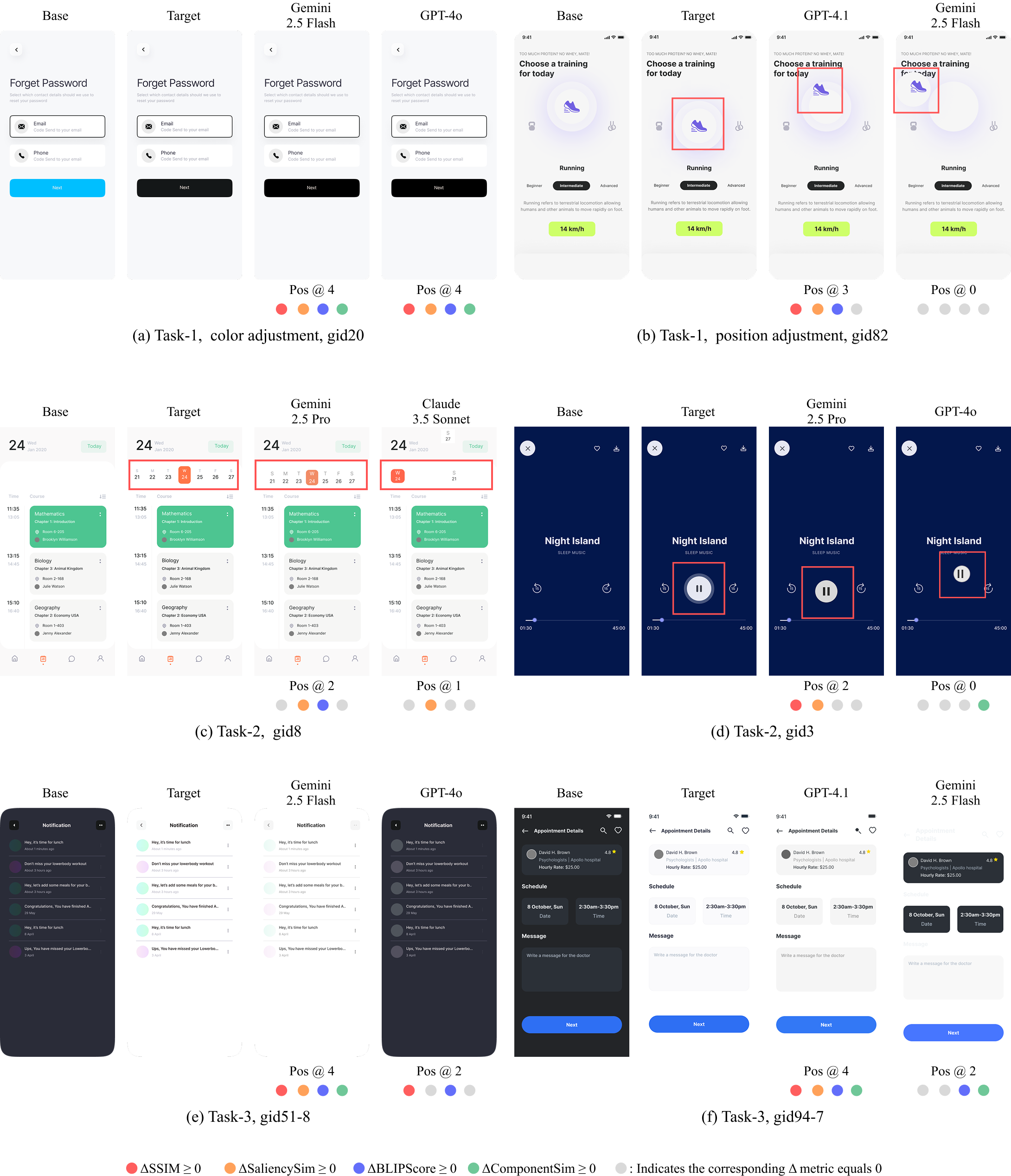}
  \caption{\textbf{Qualitative results with Pos@K annotations for the modification task.}
  Each example shows model outputs generated by editing an initial base state toward a given target design. The red bounding boxes are added for illustrative purposes and are not part of the actual UI.  Below each output, the Pos@K label indicate which of the four evaluation metrics exhibit non-negative $\Delta$ improvements compared to the target. If all metric scores remain exactly zero despite valid operations, the case is considered as Pos@0, indicating no meaningful change. Cases with no valid operation detected are reported as failure cases.
  }
  \label{fig:qual_modification}
\end{figure*}
 
\clearpage

\section{Additional Experiments}

\subsection{Task Difficulty Effect}

Table~\ref{tab:replication-type-result} shows the variation in model performance based on the task difficulty.
We define two difficulty levels based on the node count (a number of design elements) in the design: \textit{standard} (bottom 85\% percentile) and \textit{hard} (top 15\% percentile). 
Correlation analysis revealed negative associations between node count and both SSIM ($r = -0.274$, $p<0.01$) and component-wise similarity ($r = -0.349$, $p<0.01$). 
When evaluating performance separately on the hard tasks, we observed a general decrease in overall scores. 
Especially, Gemini-2.5-Pro demonstrated a stronger decrease in SSIM, whereas Gemini-2.5-Flash exhibited robustness in component-wise similarity scores.

\definecolor{bestblock}{HTML}{E8E8E8}
\renewcommand{\arraystretch}{1.1}

\begin{table}[ht]
\centering
\caption{Replication Task Scores Per Difficulty}
\label{tab:replication-type-result}
\begin{adjustbox}{max width=\columnwidth}
\begin{tabular}{@{}llcccc@{}}
\toprule
\textbf{Difficulty} & \textbf{Model} & \textbf{SSIM} & \textbf{Saliency} & \textbf{BLIP} & \textbf{Comp. Wise} \\
\midrule
\multirow{5}{*}{\makecell[l]{Standard\\(n\!=\!1258)}} & GPT-4o & 0.749 & 0.479 & 0.492 & 0.683 \\
 & GPT-4.1 & 0.776 & 0.613 & \cellcolor{bestblock}\textbf{0.649} & \cellcolor{bestblock}\textbf{0.726} \\
 & Claude-3.5-Sonnet & 0.732 & 0.482 & 0.515 & 0.679 \\
 & Gemini-2.5-Flash & 0.748 & 0.625 & 0.568 & 0.710 \\
 & Gemini-2.5-Pro & \cellcolor{bestblock}\textbf{0.788} & \cellcolor{bestblock}\textbf{0.634} & 0.621 & 0.703 \\
\midrule
\multirow{5}{*}{\makecell[l]{Hard\\(n\!=\!225)}} & GPT-4o & 0.685 & 0.472 & 0.510 & 0.613 \\
 & GPT-4.1 & \cellcolor{bestblock}\textbf{0.716} & 0.605 & \cellcolor{bestblock}\textbf{0.689} & 0.665 \\
 & Claude-3.5-Sonnet & 0.689 & 0.493 & 0.532 & 0.595 \\
 & Gemini-2.5-Flash & 0.670 & 0.583 & 0.583 & \cellcolor{bestblock}\textbf{0.665} \\
 & Gemini-2.5-Pro & 0.698 & \cellcolor{bestblock}\textbf{0.606} & 0.620 & 0.650 \\
\bottomrule
\end{tabular}
\end{adjustbox}

\begin{tablenotes}[flushleft]
\footnotesize
\item \textcolor{bestblock}{\rule{6pt}{6pt}}: the best score within \emph{each difficulty}.
\end{tablenotes}

\end{table}

\subsection{Human Preference Study}

Following the methodology outlined in Design2Code~\cite{si2024design2code}, we collect a pairwise human preference on design with design experts.
We recruited design experts through the Prolific platform, screening for participants who were fluent in English and had professional experience in relevant fields, such as UI/UX design, responsive design, or A/B testing.
The instructions provided to the participants, shown in Figure~\ref{fig:human-alignment-study-prompt}, were also closely adapted from the prior work to ensure methodological consistency.

Each annotator sees two design cases with one ground truth and labels a win, lose, or tie.
We collect 3 labels per design pair on 100 examples sampled through stratified sampling (weighted by UI type and design complexity) from our replication results. 
The final label is determined by majority voting.

\begin{figure}[H]
  \centering
  \includegraphics[width=\columnwidth]{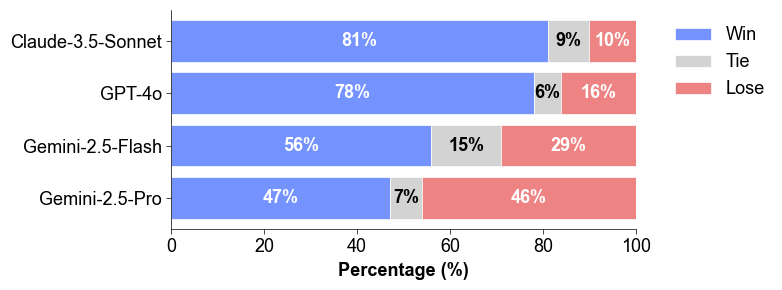}
  \caption{\textbf{Human Pairwise Preference Result}: The baseline, GPT-4.1, was majorly preferred over the other models, consistent with our metric result.}
  \label{fig:human-preference}
\end{figure}

Figure~\ref{fig:human-preference} illustrates pairwise human preferences comparing GPT-4.1 (baseline) against other models. 
Consistent with our metric results, GPT-4.1 demonstrates a strong advantage over other models. 
This inclination is especially apparent against Claude-3.5-Sonnet and GPT-4o, which showed a subpar performance in our metric result.
The results indicate that our metric well aligns with the human preference setup.

\subsection{Performance Evolution Across Turns}
We analyze how agent performance evolves across turns by visualizing turn-wise changes in four metrics.
We observe that all models exhibit measurable improvements from the initial state to the final turn, but with distinct optimization characteristics.
Notably, Claude-3.5-Sonnet shows the most stable progression, with BLIP improving from -0.008 to 0.518 and saliency increasing by +0.076, alongside low variance (SSIM std. 0.042).
GPT-4o exhibits similarly balanced improvements, including a BLIP gain of +0.505 with low saliency variability. GPT-4.1 achieves the largest overall gains, particularly in saliency (+0.204) and component-wise similarity (+0.287), though with higher variance (saliency std. 0.120).
Gemini-2.5-Flash and Gemini-2.5-Pro also show monotonic improvements, with Gemini-2.5-Pro achieving the largest SSIM gain (+0.077) and a saliency gain of +0.222, while both variants exhibit higher turn-wise variability (saliency std. 0.133). Across models, no SSIM or BLIP declines are observed in the updated results.
\begin{table}[h]
\centering
\caption{Turn-by-turn performance summary for the replication task(max 50 turns). Initial and final metric scores are shown at the first ($s_0$) and last ($s_T$) turns, respectively. $\Delta_m$ indicates the score change, and Std. denotes standard deviation over turns.}
\label{tab:turn_stats}
\begin{adjustbox}{max width=\columnwidth}
\begin{tabular}{@{}l l r r r r@{}}
\toprule
\textbf{Model} & \textbf{Metric} & \textbf{Initial} & \textbf{Final} & \textbf{$\Delta$} & \textbf{Std.} \\
\midrule
\multirow{4}{*}{GPT-4o} 
 & SSIM & 0.697 & 0.739 & 0.042 & 0.049 \\
 & Saliency & 0.408 & 0.478 & 0.070 & 0.089 \\
 & Comp.\ Wise & 0.429 & 0.671 & 0.242 & 0.097 \\
 & BLIP & -0.010 & 0.495 & 0.505 & 0.210 \\
\midrule

\multirow{4}{*}{GPT-4.1} 
 & SSIM & 0.696 & 0.767 & 0.070 & 0.054 \\
 & Saliency & 0.408 & 0.612 & 0.204 & 0.120 \\
 & Comp.\ Wise & 0.428 & 0.716 & 0.287 & 0.105 \\
 & BLIP & -0.009 & 0.655 & 0.664 & 0.265 \\
\midrule

\multirow{4}{*}{Claude-3.5-Sonnet} 
 & SSIM & 0.698 & 0.725 & 0.027 & 0.042 \\
 & Saliency & 0.407 & 0.483 & 0.076 & 0.093 \\
 & Comp.\ Wise & 0.429 & 0.666 & 0.237 & 0.084 \\
 & BLIP & -0.008 & 0.518 & 0.527 & 0.198 \\
\midrule

\multirow{4}{*}{Gemini-2.5-Flash} 
 & SSIM & 0.699 & 0.736 & 0.036 & 0.042 \\
 & Saliency & 0.408 & 0.619 & 0.210 & 0.133 \\
 & Comp.\ Wise & 0.427 & 0.702 & 0.275 & 0.104 \\
 & BLIP & -0.010 & 0.571 & 0.581 & 0.237 \\
\midrule

\multirow{4}{*}{Gemini-2.5-Pro} 
 & SSIM & 0.697 & 0.774 & 0.077 & 0.047 \\
 & Saliency & 0.408 & 0.630 & 0.222 & 0.133 \\
 & Comp.\ Wise & 0.429 & 0.694 & 0.265 & 0.088 \\
 & BLIP & -0.010 & 0.620 & 0.630 & 0.231 \\
\bottomrule
\end{tabular}
\end{adjustbox}
\end{table}

\begin{figure}[!h]
  \centering
  \includegraphics[width=\columnwidth]{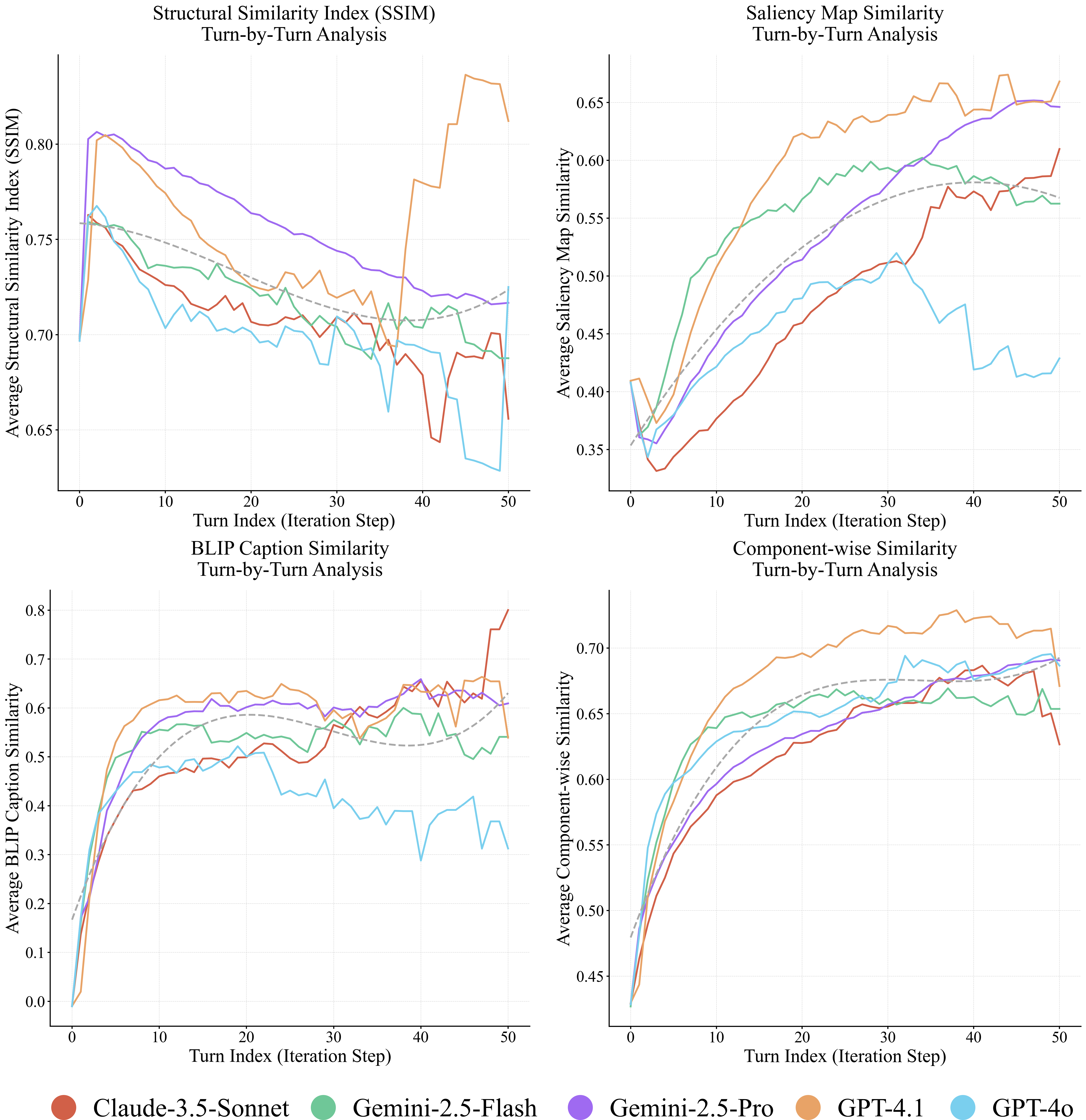}
  \caption{\textbf{Turn-by-Turn Performance Curves for Key Similarity Metrics.}}
  \label{fig:turn-metric-curve-replication}
\end{figure}

\begin{figure}[h]
  \centering
  \includegraphics[width=\columnwidth]{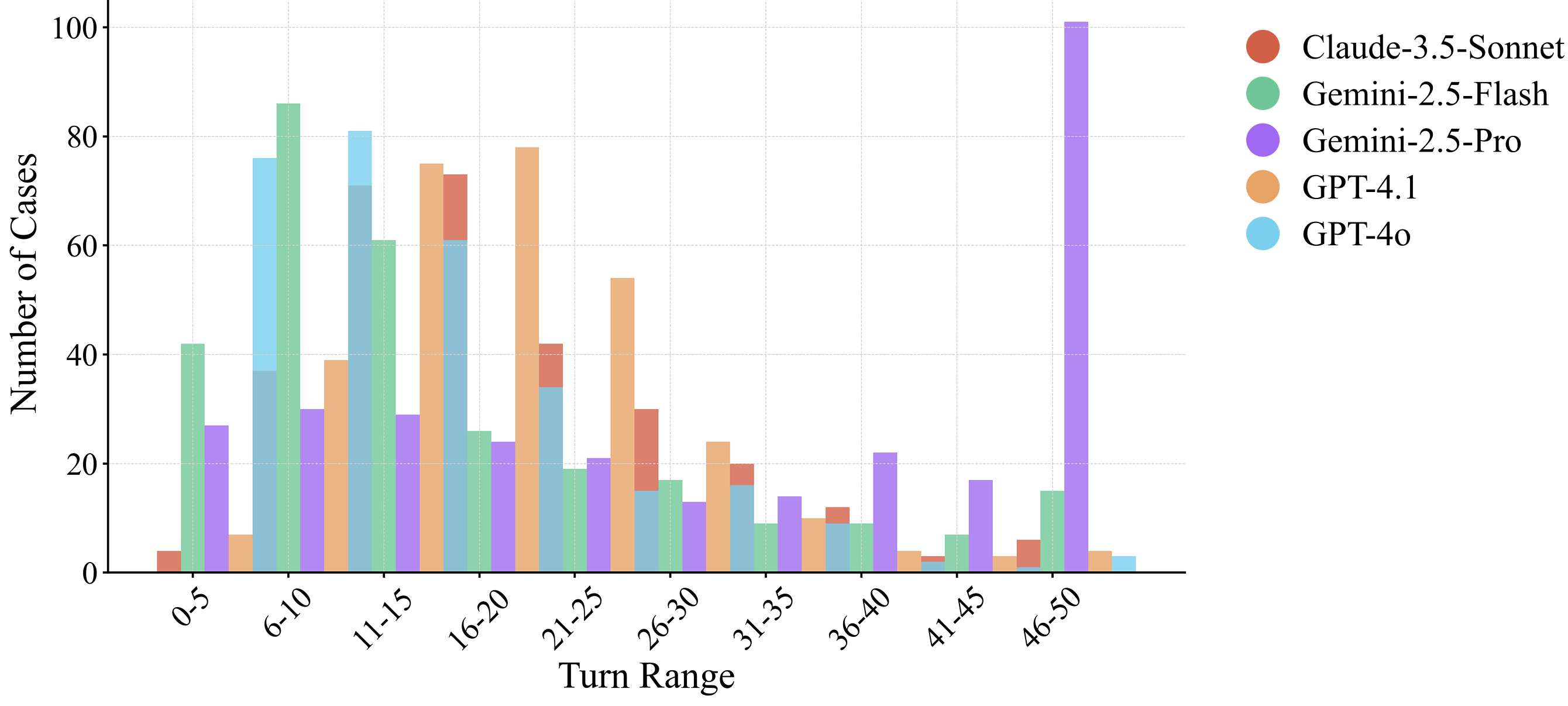}
  \caption{\textbf{Max Turn Distribution by model in replication task.}}
  \label{fig:turn-distribution-by-model-replication}
\end{figure}

These patterns reflect distinct strategic tendencies. GPT-4.1 rapidly reaches high initial scores (e.g., SSIM = 0.697) and finishes early in most cases, as shown in the turn range distribution in Figure~\ref{fig:turn-distribution-by-model-replication}, suggesting a fast-convergence strategy.
However, this often leads to performance degradation in longer tasks. Conversely, Claude-3.5-Sonnet and Gemini-2.5-Pro maintain gradual, low-variance improvements over 50 turns, indicating a more incremental optimization strategy.
SSIM shows the highest standard deviation across models, indicating potential trade-offs between structural fidelity and iterative refinement.
These findings underscore that a high final score does not necessarily indicate process stability. Models that converge to similar end states may follow markedly different turn-wise trajectories, with varying degrees of volatility and correction behavior.
Benchmarking multi-turn agents should therefore evaluate not only outcomes but also the optimization dynamics that produce them, to more faithfully capture the characteristics of real-world design workflows.

\begin{table}[t]
\centering
\small
\setlength\tabcolsep{6pt}
\caption{Tool invocation statistics per model for Task 1.}
\begin{adjustbox}{width=\columnwidth}
\begin{tabular}{lccc}
\toprule
\textbf{Model} & \textbf{Tool Precision} & \textbf{Tool Recall} & \textbf{Tool Diversity} \\
\midrule
GPT-4o               & 0.3548 & 0.9900 & 3.6100 \\
GPT-4.1              & 0.4530 & 1.0000 & 3.4000 \\

Claude-3.5-Sonnet    & 0.4080 & 0.9700 & 3.0200 \\
Gemini-2.5-Flash     & 0.5017 & 1.0000 & 2.3100 \\
Gemini-2.5-Pro       & 0.4670 & 0.9900 & 3.2700 \\
\bottomrule
\end{tabular}
\end{adjustbox}
\end{table}

\begin{table}[t]
\centering
\small
\setlength\tabcolsep{6pt}
\caption{Tool invocation statistics per model for Task 2.}
\begin{adjustbox}{width=\columnwidth}
\begin{tabular}{lccc}
\toprule
\textbf{Model} & \textbf{Tool Precision} & \textbf{Tool Recall} & \textbf{Tool Diversity} \\
\midrule
GPT-4o               & 0.5699 & 1.0000 & 5.1800 \\

\rowcolor{gray!15}%
GPT-4.1              & 0.7215 & 1.0000 & 8.3100 \\

Claude-3-5-Sonnet    & 0.7785 & 1.0000 & 3.8300 \\
Gemini-2.5-Flash     & 0.6283 & 0.9300 & 3.9900 \\

\rowcolor{gray!15}%
Gemini-2.5-Pro       & 0.5998 & 0.9000 & 5.3700 \\
\bottomrule
\end{tabular}
\end{adjustbox}
\end{table}

\begin{table}[!ht]
\centering
\small
\setlength\tabcolsep{6pt}
\caption{Tool invocation statistics per model for Task 3.}
\begin{adjustbox}{width=\columnwidth}
\begin{tabular}{lccc}
\toprule
\textbf{Model} & \textbf{Tool Precision} & \textbf{Tool Recall} & \textbf{Tool Diversity} \\
\midrule
GPT-4o               & 0.3340 & 1.0000 & 4.1400 \\
GPT-4.1              & 0.4559 & 1.0000 & 4.6200 \\
Claude-3-5-Sonnet    & 0.5168 & 1.0000 & 4.0400 \\
Gemini-2.5-Flash     & 0.6530 & 1.0000 & 3.9200 \\
Gemini-2.5-Pro       & 0.6309 & 1.0000 & 3.7800 \\
\bottomrule
\end{tabular}
\end{adjustbox}
\end{table}

\subsection{Tool Invocation Statistics}
Tool Precision (range: \([0, 1]\)) measures the proportion of model-used tools that are correct. 
It is computed as the intersection between model-used tools and human-annotated tools, divided by the total number of unique tools used by the model.
Tool Recall (range: \([0, 1]\)) measures the proportion of human-annotated tools that the model successfully used. 
It is computed as the intersection divided by the total number of human-annotated tools.
Tool Diversity represents the average number of unique tools used per case, calculated as the mean across all cases. 
Excluding the three Connection tools, all evaluation cases require at least one Inspection tool, yielding a range of \([1, 47]\).

\subsubsection{Failure Handling}
To ensure stable execution, each tool invocation is retried up to three times, and samples that fail after all retries are skipped. 
This strategy mitigates transient network or server-side errors.
We additionally observed a model-specific failure in Gemini-2.5-Flash during experiments: a \texttt{MALFORMED\_FUNCTION\_CALL} error in the Gemini API caused the model to terminate at turn~0 in 7 of 100 cases. 
These failed samples were excluded from evaluation. 
Other models did not exhibit similar API-level failures.

\begin{figure*}[h]
  \centering
  \caption{\textbf{Human Preference Study Instruction:} The instruction for the human preference study.}
  \begin{lstlisting}[
      basicstyle=\ttfamily\scriptsize,
      numberstyle=\tiny,
      frame=single,
      numbers=left,
      breakindent=0pt
    ]
TASK OVERVIEW:
In this survey, you will be given a reference mobile UI design's screenshot (Ground Truth), as well as two candidate mobile UI designs (Example A and Example B) that try to replicate the reference mobile UI design. Your task is to judge which of the two candidates is closer to the reference.

This study aims to evaluate how well AI agents can replicate reference UI designs in a canvas-based environment. Each UI design task involves a replication scenario where an AI agent is given a ground-truth UI canvas (created by a human designer) and asked to reproduce it as closely as possible by creating a new design on an empty canvas.

You will be shown screenshots of:
- The Reference Canvas (Ground Truth)
- Two Generated Canvases (produced by VLM agents attempting to replicate the reference)

INSTRUCTIONS:
1. For each question, you will see 3 images side by side:
   - Left: Baseline generation (Example A, baseline model)
   - Center: Reference image (Ground Truth)
   - Right: Comparison generation (Example B, different model)
   (A sample image is provided below for reference.)

2. Question: "Which generation is more similar to the ground truth?"

3. Choose one of the following options:
   - Left is better (Left image is more similar to reference)
   - Right is better (Right image is more similar to reference)
   - Tie (Both images are equally similar to reference)

4. COMPARISON GUIDE:

Initial Step: Content Check
- Text Content: Examine if the text on the candidate mobile UI design matches the reference. Pay special attention to missing or extra content, especially key elements like titles.
- Image Content: Assess the placement of the blue placeholder blocks (for images).
- Primary Judgment Criterion: If one example has significant missing or additional content compared to the other, it should be considered less similar to the reference.

Second Step: Layout Check
- Element Arrangement: If the content (text and images) of both examples is similarly good or bad, proceed to evaluate the arrangement of these elements. Check if their organization, order, and hierarchy match the reference.
- Secondary Judgment Criterion: If differences in layout are observed, the example with the layout most similar to the reference should be rated higher.

Final Step: Style Check
- Style Attributes: Only if Example 1 and Example 2 are comparable in content and layout, examine the style elements like font style, color, and size.
- Tertiary Judgment Criterion: In cases where content and layout are equally matched, preference should be given to the example with style attributes closer to the reference.

Overall Judgment
Based on the criteria in the order of priority (Content > Layout > Style), make an overall judgment on which example (Example 1 or Example 2) is more similar to the reference mobile UI design.

Please complete all 30 questions (6 questions per image set x 5 image sets) in this form.

Click "Next" to start the survey.
  \end{lstlisting}
  \label{fig:human-alignment-study-prompt}
\end{figure*}

\end{document}